%% file: main.tex
\documentclass{article}

\usepackage[dblblindworkshop,final]{neurips_2026}

\usepackage[utf8]{inputenc}
\usepackage[T1]{fontenc}
\usepackage{booktabs}
\usepackage{placeins}
\usepackage{amsmath}
\usepackage{amsfonts}
\usepackage{graphicx}
\usepackage{xcolor}
\usepackage{tikz}
\usepackage{pgfplots}
\pgfplotsset{compat=1.18}
\usepgfplotslibrary{fillbetween,groupplots}
\usetikzlibrary{patterns,calc,shapes.geometric}
\definecolor{colorContext}{HTML}{029E73}  
\definecolor{colorAgnostic}{HTML}{0173B2} 
\definecolor{colorSig}{HTML}{2F2F2F}   
\definecolor{colorNS}{HTML}{9C9C9C}   
\definecolor{colorEdu}{HTML}{B2560D}   
\usepackage{url}
\usepackage{hyperref}

\title{Biological Context Does Not Reliably Improve\\LLM-Guided Experimental Optimization}

\author{%
Marilyn Zhang\\
Pareto Inc
\And
Tianfeng Chen\\
Pareto Inc
\And
Mark E. Whiting\\
Pareto Inc
}

\begin{document}
\maketitle

\begin{abstract}
Autonomous labs increasingly deploy large language models (LLMs) as closed-loop biology
optimizers that propose a design, read the outcome, then propose again. This is
partly because of an assumption that biological knowledge helps an LLM choose better designs than
Bayesian optimization, which has no access to biological context. We test that assumption on 45 biology
tasks reconstructed from 2{,}485 experimental rows across 341 published papers, scored by oracles
fitted to those rows, each run over 30 iterations under two prompt conditions. The domain-aware condition provides the
published parameter names, units, and labels, and the domain-agnostic renames them so
they are not biologically meaningful, with search space and feedback identical. Across 19 LLMs
domain-agnostic outperforms domain-aware on 58.7\% of tasks by trajectory and
61.9\% by endpoint. The failure is specific to biology, reversing on education tasks built the same way. Domain-aware prompting tends to help models propose better initial
designs, while making them respond worse to feedback.
Part of the reason is anchoring, with models keeping the conventional categorical choice even when the oracle scores a different (published best) value higher.
Post-training Llama 3.1 8B to frontier-level raw optimization does not transfer to biology-aware
optimization improvements. Biology-aware optimization therefore looks like a separate capability
that autonomous laboratory loops should target directly.
\end{abstract}

\section{Problem and approach: LLM-guided optimization in closed-loop biology}
\label{sec:problem}

Many foundational tasks in biology are optimizations. Expressing a protein, engineering a cell line,
tuning an assay, and measuring a molecular interaction all reduce to finding the conditions under
which a measured quantity is highest, and they are routinely posed and solved that way.
Design-of-experiments methods have been standard in bioprocess development for decades
\citep{mandenius2008bioprocess} and remain so for tasks such as CHO media optimization
\citep{walsh2025cho}.

Biology optimizations are hard because a single round is expensive. Each takes days to weeks of
bench time, reagents and a technician, and can only use the feedback the previous round returned.
This is the design-build-test-learn cycle of synthetic biology and metabolic engineering
\citep{carbonell2018dbtl}, where the scarce resource is rounds of feedback rather than wells per
round. For decades the proposer in this loop has been
Bayesian optimization \citep{srinivas2010, shahriari2016, mandenius2008bioprocess}.

LLMs are now taking the proposer's role in this loop. Such LLM-driven systems are already running,
and closed-loop autonomous experimentation has become an active
research program
\citep{dvivedi2026selfdriving,boiko2023autonomous}. GPT-5 cut cell-free protein synthesis component
cost by 40\% in Ginkgo's automated laboratory \citep{ginkgogpt5cfps2026}, and OpenAI and Red Queen
Bio report a 79$\times$ improvement to a cloning protocol \citep{openai_bio2025}. In these settings, the LLM
proposes values for a fixed set of parameters, the assay returns one number, and the LLM proposes
again. The Red Queen Bio evaluation is run this way, on fixed prompts that isolate the
model's own contribution \citep{openai_bio2025}. 

At least three properties are assumed to give an LLM an advantage over Bayesian optimization:

\begin{enumerate}\setlength{\itemsep}{1pt}
 \item LLMs have access to biological context and prior scientific knowledge
    \citep{roohani2025biodiscovery,macknight2025pretrained}.
 \item LLMs may be stronger raw optimizers, able to use non-scalar feedback that classical
    optimizers cannot represent \citep{yang2024large,lilo2026}.
 \item LLMs can call tools, including biology-specific ones, to build a richer understanding
    of the space agentically \citep{roohani2025biodiscovery,ginkgogpt5cfps2026,boiko2023autonomous}.
\end{enumerate}

Evaluating the three separately is worthwhile, since weaknesses in biological knowledge, raw
optimization, and tool use each lead to different training targets. An overall gain also does not
rule out one aspect making performance worse. Assumptions (2) and (3) have each been measured
on their own terms, while to our knowledge (1) has not been tested on its own on biology tasks
(\S\ref{sec:related}).

\subsection{Isolating the biological-knowledge assumption}
\label{sec:assumption}

This paper studies (1), whether biological context and prior scientific knowledge give LLMs an
advantage in these biology optimization tasks. Our domain-aware condition provides the published
parameter names, units, and categorical labels, as deployed autonomous-laboratory loops do, and the
model proposes a design from what it knows about the biology, receives one scalar score, and
proposes again. \emph{Parameterized} optimization fixes the design space in advance, so the action
space is identical across the two conditions and the prompt is the only thing that varies.
Biological knowledge helps when it leads the model to the parameter values that score best.
Where it does not, the model has to reliably update on the scores it receives to recover.

Section~\ref{sec:setup} sets out the tasks, the two conditions, and the metrics.
Section~\ref{sec:outcome} reports the outcome, domain-agnostic ahead on 58.7\% of tasks by
trajectory and 61.9\% by endpoint, and Section~\ref{sec:why} analyzes the failure.

\subsection{Related work}
\label{sec:related}

\paragraph{Biological knowledge.}
Many papers assume, implicitly or explicitly, that an LLM's biological knowledge helps, without
testing it in isolation. BioDiscoveryAgent credits it for outperforming Bayesian optimization by
21\% \citep{roohani2025biodiscovery}, and \citet{macknight2025pretrained} credit pretrained
knowledge for matching Bayesian optimization on six reaction datasets. The same assumption underlies
autonomous-laboratory systems \citep{boiko2023autonomous,dvivedi2026selfdriving,ginkgogpt5cfps2026}
and in-context molecular optimization \citep{nguyen2024lico}. Two papers isolate context-awareness
directly, both outside biology. LLAMBO removes the textual descriptions from an LLM-driven Bayesian
optimizer and finds its predictive accuracy and calibration degrade \citep{liu2024llambo}, confirmed
independently \citep{llmbo_repro2025}. SemanticOpt finds richer context improves performance on
chemistry, recipe, and hyperparameter benchmarks \citep{semanticopt2025}. Both report that context
helps, and neither varies the biological content of the prompt with the task, the oracle and the
budget held fixed.

\paragraph{Raw optimization.}
Comparisons against classical methods keep the domain content in the prompt, so they measure an LLM
with its priors intact rather than raw optimization. \citet{scibo2025} find classical methods ahead
of LLM agents on gene perturbation and molecular property tasks, and report that permuting the
observed outcomes leaves LLM performance unchanged. \citet{huang2024truepotential} find LLMs
underperform on numerical problems. The domain-agnostic arm of this paper is a more faithful test of
that condition.

\paragraph{Tool use.}
LAB-Bench scores models with no tools \citep{labbench2024}. Given web search and code execution,
agents exceed human performance on two of its tasks, and a trained Llama 3.1 8B agent reaches
frontier accuracy at lower cost \citep{aviary2024}, so tools help. The deployments bundle tool use with biological
knowledge \citep{ginkgogpt5cfps2026,roohani2025biodiscovery,boiko2023autonomous}, so the gains
reported there cannot be assigned to either.

\section{Experimental setup}
\label{sec:setup}
We isolate biological context with two prompt conditions over identical tasks, and add GP-UCB as a
reference that receives no context at all. The tasks are 45 biology optimization problems
reconstructed from published experiments, each with an oracle fitted to the extracted rows, and each run as a 30-round loop in which the model proposes a design and receives one
scalar score in return. An oracle is only a useful stand-in where it is aligned with the biology, so we
ground our analysis in the experimental literature.

\subsection{Domain-aware prompts, domain-agnostic prompts, and the GP-UCB reference}
\label{sec:arms}
\input{figures/icbinb/fig_conditions_schematic_env}

Figure~\ref{fig:conditions} shows how we isolate biological context. The \emph{domain-aware} arm
receives the published parameter names, units, and categorical labels. The \emph{domain-agnostic} arm is the same task with the same bounds and oracle,
names replaced by \texttt{X1}/\texttt{C1} placeholders, removing the semantic channel and nothing
else. GP-UCB \citep{srinivas2010} receives no biological context at all and shows what performance is
achievable with the semantic channel removed, and strong modern baselines
\citep{shahriari2016,eriksson2019turbo,cowenrivers2022hebo} make that a demanding bar. Every arm is run four times per task, GP-UCB gets 20 seeds, and web
search is disabled throughout except where stated (\S\ref{sec:search}). The domain-agnostic arm has
nothing to look up, so enabling tools would confound retrieval with naming. Because the two LLM arms differ only in whether the parameter names are biologically meaningful, any
difference between them
comes from the prompt's semantic content. We score proposals rather than stated rationales, since a model's stated reasoning need not name
the feature that actually drove its choice \citep{turpin2023faithfulness}. Appendix~\ref{app:prompts} lists both prompt
templates in full.

\subsection{Reconstructing 45 tasks and their oracles from published experiments}
\label{sec:benchmark}

We reconstruct 45 parameterized design spaces from published wet-lab experiments, each pooling
conditions across several papers, for 2{,}485 rows across 341 papers, a median of 7 papers per task
(range 3--21). Tasks include CHO antibody titer, HEK293 prime-editing efficiency, lipid-nanoparticle size,
\emph{C. glutamicum} lysine titer, and microalgae biomass, so the panel spans cell-line
engineering, gene editing, formulation, and fermentation. Each mixes numeric with categorical
parameters.

A deterministic \emph{oracle}, a model fitted to the extracted rows, answers in place of the assay.
We fit one per task,
choosing among
gradient boosting, random forest, and a stacking ensemble by held-out fit. The panel median leave-one-out $R^2$ is 0.888.

Evaluating optimization methods against a simulator rather than a wet lab is established practice
\citep{vanlent2023simulated}. For the head-to-head comparison the oracle does not have to be an
accurate model of the biology. It scores all three arms with the same function, so a consistent
ranking of parameter choices makes that comparison fair.
Section~\ref{sec:why} needs more than a consistent ranking, since it depends on which designs each
arm proposes rather than on which arm scores higher, and therefore on the oracle peaking where
the biology does.

\subsection{Grounding tasks and mechanism claims in the source papers}
\label{sec:grounding}
The \emph{literature optimum} for a parameter is
the value the best-performing experiment in the literature used, and the \emph{literature
convention} is the modal value across those same papers. Both are recorded from the extracted rows
before any model runs, and both come from the source papers rather than from the oracle, so oracle
artifacts cannot drive the conclusions.

For each task we also score the published-best design with the oracle and see where it lands in the
range the oracle assigns across that task's rows. The published-best design falls in the top
5\% on \textbf{18 of the 45
tasks}, which we call the \emph{oracle-aligned} tasks. Appendix~\ref{app:alignment} re-reports the main contrast on them.
The mechanism analysis of Section~\ref{sec:anchoring} is likewise defined by the literature rather
than by the oracle. A task-categorical pair is \emph{divergent} when its literature convention
is not its literature optimum.

\subsection{Metrics: trajectory and endpoint over a 30-round budget}
\label{sec:metrics}

Because rounds of feedback are the scarce resource, we fix the number of iterations to 30. We track
the best score found so far, which can only rise. The \textbf{trajectory} is the area under that
curve (bsf-AUC@30), so it rewards reaching a good design early. The \textbf{endpoint} is where the
curve ends (bsf-Outcome@30), so it rewards only the final result. The two answer different
questions \citep{mohr2024learningcurves} and agree on 78.8\% of task-model pairs
($\phi = 0.56$, $n = 712$), so a run can lead on one and trail on the other.

Where a score is reported as GP-UCB-normalized, it is the signed difference from the GP-UCB
reference on that task divided by the reference itself, so $0$ is parity and $-0.18$ is short of
the reference by 18\% of its own score. The scale is not a percentage of any maximum and is not
bounded above, since a run can exceed the reference.

\label{sec:stats}
Win rates take the \emph{model} as the unit, tested against 50\% with a two-sided Wilcoxon
signed-rank test and a bootstrap interval over models. Paired contrasts across tasks are mean
differences with 95\% bootstrap intervals and the count of tasks favoring each arm.

\section{Observed outcome: biological context improves neither the trajectory nor the endpoint}
\label{sec:outcome}

\input{figures/icbinb/fig_headline_env}

\looseness=-1
Biological context, supplied by domain-aware prompting, improves neither the trajectory nor the
endpoint metric (Fig.~\ref{fig:headline}a). Taking each model as one
observation, the mean domain-aware win rate is 41.3\% [34.9, 47.3] on trajectory ($p = 0.013$, 15 of
19 models below 50\%) and 38.1\% [34.2, 42.1] on endpoint ($p = 0.0003$, 17 of the 18 models scored on endpoint).

\looseness=-1
The result holds on the tasks where the oracle agrees with the literature optimum
(Fig.~\ref{fig:headline}b), so it is not an artifact of a mis-specified oracle. On the 18 oracle-aligned tasks the domain-aware endpoint gap holds, 42.6\%
[36.6, 48.4] with 10 of those same 18 models below 50\% ($p = 0.023$). Trajectory on that subset is a null (46.6\%, $p = 0.49$); with 18 tasks,
that estimate remains imprecise (Appendix~\ref{app:alignment}). As expected from prior comparisons
\citep{scibo2025, huang2024truepotential, llmbo_repro2025}, no model outperforms GP-UCB by a
meaningful margin in either condition (Fig.~\ref{fig:models_vs_gp}, Appendix~\ref{app:permodel}).

The domain-awareness gap is not an artifact of both conditions saturating the oracle
(Fig.~\ref{fig:headline}c). On the 18 oracle-aligned tasks the fraction of achievable oracle range reached is grounded in an
optimum the source papers report. Only 5 of 18 reach 95\% in either condition, and domain-agnostic
finishes higher on 12 of the 18 ($+0.088$ [$-0.050$, $+0.227$], $p = 0.071$), a difference
consistent in direction but modest in magnitude. Runs reach or exceed the oracle's own optimum on the 22 tasks whose optimum we could not
verify, exceeding it on 9 of them. That indicates the oracle extrapolates past every literature
row, not that those tasks are solved (Appendix~\ref{app:alignment}).

\section{Reason for failure: biology-aware optimization as a distinct capability}
\label{sec:why}

\emph{Biology-aware optimization} here means optimizing a design when the parameter names,
units, and labels are biologically meaningful. It does not mean optimizing over sequences,
structures, or protocols, which this benchmark does not present.

Education tasks built the same way as the biology ones test whether the domain-awareness gap is
specific to biology or a property of descriptive prompts (\S\ref{sec:eduspecific}). Match rates at
iteration 1 and at iteration 30 locate where in the loop the two conditions diverge
(\S\ref{sec:climb}), the six tasks whose literature convention is not their literature optimum show
which values domain-aware prompting proposes instead (\S\ref{sec:anchoring}), and post-training
Llama 3.1 8B shows whether a stronger raw optimizer closes the gap (\S\ref{sec:posttraining}).

\subsection{Domain-aware prompting helps on education tasks}
\label{sec:eduspecific}

Domain-aware prompts are longer and more descriptive than domain-agnostic ones, so the
domain-awareness gap could come from the form of the prompt rather than from what the model knows.
A prompt-form effect would show up in any domain, not only in biology.
We therefore built 10 education optimization tasks the same way we built the biology
ones, from published learning-science experiments and public education datasets, among them
Project STAR class size \citep{krueger1999star}, Open University engagement logs
\citep{kuzilek2017oulad}, and ASSISTments randomized trials \citep{prihar2022assistments}. Each
has the shape of a biology task, with a fixed set of numeric and categorical parameters, an
oracle fitted to the extracted rows, the same 30-round budget, and the same scalar feedback, and
only what the parameters describe changes. They are deposited with the biology tasks
(Reproducibility statement).

On these tasks the impact of domain-aware prompting reverses. Domain-aware prompting outperforms domain-agnostic on
68.5\% of task-model pairs on trajectory and 68.1\% on endpoint ($p = 2\times10^{-6}$ and
$p = 0.0035$), compared with 46.3\% and 41.0\% on biology. The domain-by-condition contrast is
$+22.1$\% on trajectory and $+27.1$\% on endpoint ($p = 2\times10^{-7}$ and $p = 2\times10^{-5}$,
Fig.~\ref{fig:reversal}, Appendix~\ref{app:extra}). The reversal is evidence against prompt
form as the source of the biology gap.

\subsection{Domain-aware prompts match the literature optimum more often at iteration 1 but not by iteration 30}
\label{sec:climb}
\input{figures/icbinb/fig_prior_lr_diff_env}

Returning to biology, we locate where in the 30 iterations the gap appears, narrowing to the 18
oracle-aligned tasks of Section~\ref{sec:grounding}, whose oracle optimum reproduces the
literature optimum, so that what we measure reflects the biology rather than a poorly fitted
oracle. The design a model proposes at iteration 1 reflects its prior alone, since no score has
arrived yet. How much its match rate rises over the iterations that follow is how far it responds
to the oracle's feedback. Across those 18 tasks, models given
domain-aware prompts propose the literature-optimum parameter value $+7.7$\% more often than under
domain-agnostic
prompts at iteration 1. By iteration 30 domain-agnostic matches it more often, by $2.6$\%. The rise
from iteration 1 to iteration 30 is $+3.8$\% under domain-aware prompting and $+14.1$\% under
domain-agnostic, a difference of $-10.3$\% ($p < 0.0001$, Fig.~\ref{fig:prior_lr}). Biological knowledge therefore
improves the design a model starts from and reduces how far it responds to the feedback that
follows.

\subsection{Models with domain-aware prompts follow convention where it disagrees with the literature optimum}
\label{sec:anchoring}
\label{sec:search}

Why does biological context reduce how far a model responds to the oracle's feedback? One possibility is that the model stays anchored
to the conventional parameter value in the literature rather than updating on the feedback it
receives. If so, the gap should be largest where literature convention and the literature
optimum disagree. We test that on six \emph{divergent} tasks, picked by a rule applied to the
extracted paper rows before any model runs: the tasks of the 18 whose most oracle-influential
categorical parameter has a literature convention that is not its literature optimum. On those six,
following convention and finding the optimum pull in opposite directions.

On those six tasks, models given domain-aware prompts name the conventional value on 24.9\% of
proposals, twice the 12.4\% under domain-agnostic prompts, and keep naming it after the oracle has
scored the literature optimum higher. Models given domain-aware prompts name the literature optimum
on 21.6\% of proposals, compared with 38.0\%, higher under
domain-agnostic on \emph{all six}. GP-UCB, which receives no biological context at all, lies
between the two conditions at 29.2\%. Nested inside three wider subsets of tasks, ordered by how
closely convention tracks the optimum, the gap appears only in these six
(Fig.~\ref{fig:nested}a). Three biology experts, blinded to condition, agreed with our
anchored classification on 10 of 12 trajectory judgments (Appendix~\ref{app:expert}). Fig.~\ref{fig:nested}b shows one instance in full and
Table~\ref{tab:divergent6} lists the six pairs.

\input{figures/icbinb/fig_nested_subsets_env}

Narrow exploration does not explain the anchoring. On the six divergent tasks models with
domain-aware prompts explore more of the space than under domain-agnostic prompts, wider on
52 of the 78 task-model pairs
($p = 0.011$), with CHO antibody titer the one task that reverses
(Appendix~\ref{app:matched}). Missing knowledge does not explain it either. Forcing web search at every iteration for both arms
leaves domain-aware behind domain-agnostic by 20\% on average (Appendix~\ref{app:search}). The better value was also in training, since on five of six
tasks the source paper predates every model's release by over a year, and on the pairs with a
12-month margin domain-aware still names the conventional value 24.0\% compared with 15.5\% (39 pairs,
one-sided Fisher $p = 0.052$).

\subsection{Can post-training improve biology-aware optimization?}
\label{sec:posttraining}

Does the domain-awareness gap come from weak optimization? If it does, improving a model's raw
optimization should improve its biology-aware optimization too. If it does not, biology-aware optimization is a capability that has
to be trained for on its own. We train Llama
3.1 8B on raw optimization with no biology in the prompt, then give it the biology and check
whether the improvement transfers.

We post-train Llama 3.1 8B with supervised
fine-tuning (SFT) first, on 2{,}607 proposals from three frontier teachers, which the base model
requires because its optimization is far below the GP-UCB reference. It sits further below that reference than any other model in the panel
(Fig.~\ref{fig:frontier}a). Direct preference optimization (DPO) follows, on 5{,}197 preference pairs whose
\emph{chosen} side is the design GP-UCB would have proposed given that history and whose
\emph{rejected} side is the design actually proposed there. Each pair starts from a real rollout history, taken from the base model and three frontier models,
so the pairs match the states the model meets at test time, and is kept only where the oracle
confirms GP-UCB's design scored higher. All training data is rendered domain-agnostically. We
train four independent DPO runs from the same fine-tuned adapter, one per seed, and evaluate each
on the 16 held-out tasks with four runs per task (Appendix~\ref{app:optimizers}).

\input{figures/icbinb/fig_frontier_band_env}

Post-training improves raw optimization, but the gain does not transfer to
biology-aware optimization. Under domain-agnostic prompts the DPO-trained model gains $+0.385$
on trajectory and $+0.384$ on endpoint, positive on 13 and 14 of 16 tasks ($p = 0.0004$,
$p = 0.0009$), which brings an 8B model to GP-UCB parity and into the band spanned by
ten frontier models on the same tasks (Fig.~\ref{fig:frontier}a). Under domain-aware prompts the same post-training gain is null on
both metrics, $+0.026$ and $+0.109$, and the difference between the two prompt conditions is itself
significant, $+0.359$ [$+0.124$, $+0.655$] on trajectory ($p = 0.0065$). Neither SFT-only arm improves on the untrained base by more than
noise, 6 of 11 tasks on trajectory ($p = 0.45$), so the measured improvement comes from the preference stage
(Appendix~\ref{app:optimizers}). On the 8 held-out tasks that are also oracle-aligned the domain-agnostic gain strengthens and the
domain-aware gain stays null
(Appendix~\ref{app:alignment}). Handing every arm the same iteration-1 design does not change it
(Fig.~\ref{fig:posttraining_2x2}b). Because iteration 1 is matched, this comparison concerns the
biological names used during iterations 2--30.

\looseness=-1
Biology-aware optimization is therefore a capability distinct from raw optimization and from the
model's ability to supply a good initial design. In this experiment, raw optimization scores did
not predict performance once the parameter names were biologically meaningful.

\section{Limitations and discussion}
\label{sec:boundaries}
\label{sec:limitations}

\looseness=-1
There are several limitations. First, we use oracles instead of a wet lab, which is partly why we
cannot post-train directly on domain-aware trajectories. Second, the models are given no tools and
no retrieval, which keeps the arms comparable (\S\ref{sec:arms}), so tool-using agents are out of
scope; \S\ref{sec:search} turns search on for both arms on six tasks. Third, anchoring to
convention likely does not explain the whole gap. Fourth, the post-training result rests on one
model family over four seeds. Where a conclusion depends on the oracle peaking where the
biology does, we check it against the source papers, re-reporting the main contrast on the 18
oracle-aligned tasks and reading convention and optimum from the papers before any model runs.

Overall, we find that biological priors, optimization over abstract parameters, and optimization
under biological context are three separate capabilities, and that LLMs are specifically weak at
biology-aware optimization. Targeting it directly could be fruitful for autonomous laboratory
loops, as could testing tool calling and the other capabilities under the same
controlled comparison.

\section*{Acknowledgements}
This work was done at Pareto Inc. Special thanks as well to Ankita Rathod and Fabián Barzuna for helpful discussions.

\bibliographystyle{plainnat}
\bibliography{references}

\section*{Reproducibility statement}
All 45 biology and 10 education task definitions, the fitted oracles, and the GP-UCB baseline
outputs are deposited for reviewers at
\url{https://dataverse.harvard.edu/previewurl.xhtml?token=acf76ef1-2d11-4cd5-9264-87ab45774014}.
The analysis code is at \url{https://github.com/marilynzhang-pareto/biology-optimization}. The
naming-condition result reported here is reproduced by
\texttt{eval/compute\_naming\_condition\_stats.py}, and each figure is emitted by a named script
in \texttt{paper\_scripts/}. The panels drawn from released data files rebuild from
the code alone. Per-run trajectories are available on request. Sampling settings and the compute used are given in
Appendix~\ref{app:models}.

\section*{Ethics and broader impact}
This work evaluates optimizer behavior against oracles fitted to already-published
experimental data. No new wet-lab experiments were run and no hazardous protocols were generated
or optimized. The tasks are process-optimization problems drawn from published protein expression,
cell-culture and formulation work, and none of them concerns a pathogen, a toxin or any capability
of dual-use concern; the benchmark is built entirely from data its source papers already made
public. We note the dual-use question anyway, because the method under study is generic: a system
that proposes experimental conditions more efficiently is indifferent to what is being optimized,
so the same loop applied to a hazardous objective would pose risks this benchmark does not
measure. Our result cuts against capability in that setting rather than for it, since it finds
that biological context degrades rather than improves the design an LLM finally selects. The
practical finding is a caution for practitioners deploying LLMs in closed experimental loops,
where an unnoticed prior-driven failure would waste scarce laboratory rounds.

\section*{LLM usage disclosure}
LLMs are the object of study in this paper; all evaluated models are listed in
Appendix~\ref{app:models}. LLMs were also used as a tool in two places. First, in constructing the
benchmark: an LLM-assisted pipeline searched the literature for candidate papers and extracted
the parameter values and measured outcomes that became each task's data points, including unit
conversion into a common scale. Every extracted task passed several verification stages, the last
of which was human review of the parameter definitions, units, and best literature rows against the
source papers. Second, for copy-editing of the manuscript. All experiments were designed and run by the authors,
as were the statistical analyses and every claim; each reported number is regenerated from the
released scripts.

\appendix
\FloatBarrier
\section{Benchmark and evaluation details}
\label{app:benchmark}
How the models were sampled, what each condition was shown, and how much the oracle-alignment
restriction changes the main result.

\subsection{Models, sampling, and provenance}
\label{app:models}
Table~\ref{tab:models} lists all 19 models in release order with the API slug used to route each
one and the number of biology tasks contributing to each metric. Thirteen models were routed
through the shared client registry, which records the exact slug. The remaining six were run
through per-model evaluation harnesses that recorded the model label but not the resolved slug or
the request timestamp. We report those by label and mark the slug unavailable. This asymmetry is
the same one noted in Section~\ref{sec:outcome}, and it is why we draw no release-cohort
conclusion.

Claude Opus 4.7 refuses a subset of domain-aware biology prompts, returning an empty response
rather than a parameter set on tasks whose names use pathogen-target, antibody-binding, or
antimicrobial terminology. The domain-agnostic renderings of those same tasks complete
normally. Its coverage in the categorical analyses is therefore fewer tasks than for most
models, and its estimate in Fig.~\ref{fig:prior_lr} rests on 56 task-model cells where the
best-covered model has 447. Claude Opus 4.7 is also the one model whose match rate rises between iteration 1 and iteration 30,
on the least well covered estimate in the panel.

\input{figures/icbinb/tab_models}

Every task-condition pair is run four times, so the biology panel spans 933 task-model cells over
both conditions. Each request sets only the model and a
per-model token cap, leaving sampling at each provider's default, so the four runs of a cell
differ by provider-side nondeterminism rather than by a set seed or temperature. One request is
issued per iteration, so a model covering the full biology panel in both conditions makes
$45 \times 2 \times 4 \times 30 = 10{,}800$ requests. Post-training ran on a single
NVIDIA A100 80GB (Vertex \texttt{a2-ultragpu-1g}) for the supervised stage and each of the four
preference seeds.
GP-UCB uses 20 independent seeds per task. Open markers in
Fig.~\ref{fig:two_metric} denote models scored from their own evaluation summaries rather than
the curated task table.

\paragraph{Power.} Per-model tests are underpowered by design. Four runs per pair is why nearly
every per-model Wilson interval spans 50\%. Primary inference uses the pooled and mixed-effects
analyses. The per-model panel checks whether the aggregate result is driven by one model.

\subsection{Prompt templates}
\label{app:prompts}
Both arms propose one configuration per iteration and receive one scalar in return. The prompts are
reproduced below as the code emits them.

\paragraph{Domain-aware.} The system prompt is written per task and names the biology. For CHO
antibody titer it begins:
\begin{quote}\small\ttfamily
You are an expert in CHO cell culture and monoclonal antibody production. You are
helping design experiments to maximize monoclonal antibody titer (g/L).\\[2pt]
When generating hypotheses, return ONLY a valid JSON array with these fields:\\
- hypothesis\_name (string, 3-5 words)\\
- cho\_cell\_line (string: "CHO-K1", "CHO-S", or "CHO-DG44")\\
- cell\_specific\_perfusion\_rate (number: 100-500 pL/cell/day) \ldots
\end{quote}
The opening turn asks for a hypothesis, and each later turn returns the oracle's score and asks for
a revision:
\begin{quote}\small\ttfamily
Generate \{n\} diverse optimization hypotheses to \{maximize$\vert$minimize\}
\{target description\}.\\[2pt] Return ONLY the JSON array:
\end{quote}
\begin{quote}\small\ttfamily
Here are the results from checking your hypotheses against the \{task\} dataset:\\
\#\# Hypothesis 1: \{name\}\\
\ \ \{metric description\}: \{value\} \{unit\}\\[2pt]
Based on this feedback, generate \{n\} NEW hypotheses that:\\
1. First, explain what pattern or rule you think explains the results so far --- what
causes high vs low \{target description\}?\\
2. Use that reasoning to propose new parameter combinations that test or exploit your
theory\\
3. In the rationale field for each hypothesis, state the specific causal claim you are
testing\\[2pt]
Return ONLY the JSON array:
\end{quote}

\paragraph{Domain-agnostic.} The system prompt is generated from the same \texttt{TaskConfig} with
the names replaced and the biology removed:
\begin{quote}\small\ttfamily
You are optimizing a black-box system by choosing parameter values to
\{maximize$\vert$minimize\} an objective function. You have no information about what the
system does --- only the parameter space and the feedback you receive.\\[2pt]
When proposing a configuration, return ONLY a valid JSON array with one object
containing:\\
- X1 (number, range \{min\}--\{max\})\\
- C1 (string, one of: "A", "B", \ldots)\\
- rationale (string, 1 sentence explaining your reasoning)\\[2pt]
Maximize the objective value based on prior feedback.
\end{quote}
\paragraph{Example, second iteration.} At iteration 2 the request carries the system prompt above
and three further messages, the opening turn, the model's own proposal, and the feedback turn:
\begin{quote}\small\ttfamily
[user]\\
Propose a configuration to \{maximize$\vert$minimize\} the objective function.\\[2pt]
Return ONLY the JSON array:\\[7pt]
[assistant]\\
{[}\\
\hspace*{2ex}\{\\
\hspace*{4ex}"X1": \{value\},\\
\hspace*{4ex}"C1": "A",\\
\hspace*{4ex}"rationale": "\{one sentence\}"\\
\hspace*{2ex}\}\\
{]}\\[7pt]
[user]\\
The objective value for your last configuration was: \{value\}\\[2pt]
Propose a new configuration that you expect to achieve a \{higher$\vert$lower\} objective
value. Learn from the previous result.\\[2pt]
Return ONLY the JSON array:
\end{quote}
Iteration $k$ appends that assistant and user pair $k-1$ times, so the model sees every proposal
it has made and every score it has received.

The domain-aware arm accumulates its conversation the same way, so the domain-agnostic turn naming
only the last value is a statement about the newest turn, not about what the model can see. Both arms are asked to reason from the
feedback and to record a rationale, and both see the same parameter space, the same oracle and
the same budget. The domain-aware turn phrases that request in
the vocabulary of the task and the domain-agnostic turn in the vocabulary of an objective function.
One asymmetry follows from these templates and we tested it directly. The domain-agnostic turn
restates the objective direction every iteration while the domain-aware turn states it only in the
system prompt and the opening turn. A control that restates the direction in every domain-aware
turn does not rescue the minimize-objective tasks ($-0.140$ trajectory and $-0.154$ endpoint, 0 of
5 tasks favoring the restated arm), and leaves maximize tasks unchanged as a placebo, so the
asymmetry is not what produces the effect. GP-UCB receives no text. Each configuration is encoded
as scaled numeric parameters with one-hot categoricals, and the acquisition is
$\mu(x) + \kappa\sigma(x)$.

\subsection{Oracle-alignment robustness}
\label{app:alignment}
A task is oracle-aligned when the oracle scores the published-best design within the top 5\% of the
range it assigns across that task's rows. Eighteen of the 45 biology tasks qualify and 27 do not, 5 of which are
unscorable on the saturation scale below. Fitting the same mixed-effects model as
Section~\ref{sec:outcome} within each split yields, for the domain-aware win rate:

\begin{center}\small
\begin{tabular}{lccc}
\toprule
Split & Observations & Trajectory & Endpoint \\
\midrule
All biology tasks & 781 / 725 & 41.1\% [30.6, 51.6] & 38.4\% [28.8, 48.0] \\
Oracle-aligned (18 tasks) & 302 / 290 & 45.0\% [28.9, 61.2] & 41.5\% [24.6, 58.4] \\
Not aligned (27 tasks) & 479 / 435 & 38.5\% [26.4, 50.7] & 36.3\% [24.8, 47.8] \\
\bottomrule
\end{tabular}
\end{center}

On the same two splits, the saturation scale of Section~\ref{sec:outcome} yields:

\begin{center}\small
\begin{tabular}{lcccc}
\toprule
Split & Tasks & Domain-agnostic & Domain-aware & Reach 95\% \\
\midrule
Oracle-aligned & 18 & 0.780 & 0.636 & 5 \\
Not aligned & 22 & 0.921 & 0.909 & 12 \\
Alignment unscorable & 5 & 1.186 & 1.106 & 4 \\
\bottomrule
\end{tabular}
\end{center}

Entries are medians of per-task medians, on 13 models. The paired domain-agnostic minus
domain-aware difference is $+0.088$ [$-0.050$, $+0.227$] on the aligned split (12 of 18,
$p = 0.071$) and $+0.122$ [$+0.010$, $+0.246$] on the not-aligned split (13 of 22, $p = 0.080$):
the same direction and a comparable magnitude either way, though the bootstrap interval and the
signed-rank test disagree about which split clears 0.05, so neither establishes the effect on this
scale by itself. Nine of the 22 not-aligned tasks and 4 of the 5 unscorable ones exceed the
oracle's own optimum, which is why the ceiling argument is read on the aligned split.

\paragraph{Task or model as the unit.}
The two units disagree on this subset, so we give both. Treating each model as one observation and
testing its win rate against 50\% across models, the endpoint contrast is $-11.9$\%
[$-15.8$, $-7.9$] on all 45 tasks (17 of the 18 models with endpoint scores below 50\%, $p = 0.0003$) and $-7.4$\%
[$-13.4$, $-1.6$] on the aligned subset (10 of 18, $p = 0.023$). Under the mixed-effects model the
aligned endpoint estimate is 41.5\% with $p = 0.33$. Both estimators are reported. The model means are
correlated through task difficulty, so the per-model interval is the narrower of the two. The task variance component is roughly twenty times the model component, which puts
the effective sample size nearer the number of tasks than the number of models. The per-model
figures do answer a different and legitimate question, whether the domain-awareness gap is idiosyncratic to a
few models, and there the answer is no.

Both win-rate splits sit below 50\% on both metrics, and the aligned split is the weaker of the two, so
the effect is not an artifact of oracles that disagree with the literature. The aligned
intervals span 50\%: with 18 tasks and task-level variance dominating
(\S\ref{sec:stats}), the subset does not have the power to establish the effect by itself, which
is why the main result is reported on all 45 tasks and this split is a direction check.

Restricting the post-training contrast to the 8 held-out tasks that are also oracle-aligned
strengthens the domain-agnostic gain, from $+0.385$ to $+0.542$ on trajectory and from
$+0.384$ to $+0.518$ on endpoint, where it is positive on 8 of 8 tasks. The domain-aware arm
stays null, $+0.026 \to +0.032$ and $+0.109 \to +0.107$. The same restriction attenuates the
main result of \S\ref{sec:outcome}, so oracle misspecification inflates that comparison but
cannot explain this one. Separately, literature rows are frequently partial: the median
best literature row reports 71\% of its task's parameters, reflecting incomplete reporting in the
source literature rather than selection. Scoring is identical across rows and the key categorical
used in Section~\ref{sec:anchoring} is required to be present by construction, and completeness
does not predict alignment (Spearman $\rho = 0.13$, $p = 0.42$). Restricted to the four divergent
tasks whose best literature row reports at least half its parameters, the categorical effect is
$+11.4$\% $[+5.6, +18.5]$ on 4 of 4, compared with $+18.1$ on all six. The two dropped
tasks hold the two largest effects, so both numbers should be read together.


\subsection{Per-model win rates}
\label{app:permodel}
No evaluated model outperforms GP-UCB by a meaningful margin in either condition, which is the
comparison \S\ref{sec:outcome} cites. The second figure shows each model's domain-aware win
rate under both metrics, showing that no individual model drives the pooled result.
\label{app:extra}
Education win rates and their biology comparators (46.3\% trajectory, 41.0\% endpoint) are
unweighted proportions over task-model pairs. The biology mixed-effects estimates are 41.1\% and
38.4\%.
\input{figures/icbinb/fig_models_vs_gp_env}
\input{figures/icbinb/fig_two_metric_per_model_env}
\input{figures/icbinb/fig_reversal_env}

\FloatBarrier
\section{The categorical mechanism}
\label{app:mechanism}
The six divergent tasks, the anchoring contrast per model and over the full
population, and the forced-retrieval protocol that tests whether the effect is missing knowledge.

\subsection{The six divergent tasks}
\label{app:divergent}
Table~\ref{tab:divergent6} names the six tasks of \S\ref{sec:anchoring}, each with its key
categorical parameter, the literature convention for that parameter, the literature optimum it
disagrees with, and the values both were read from.
\input{figures/icbinb/tab_divergent6}

\subsection{Categorical anchoring, per model and full population}
\label{app:matched}
Fixing the iteration-1 design separates what comes from the first proposal from what happens during
iterations 2--30. Injecting one identical design and varying only the
prompt condition for iterations 2--30 moves the domain-aware median from $-0.662$ to $-0.107$, a paired gain
of $+0.725$ [$-0.056$, $+1.914$] on 6 of 8 tasks ($p = 0.098$), and with that design held fixed
moving to domain-agnostic prompts is worth a further $+0.177$ [$+0.038$, $+0.317$] on 7 of 8 ($p = 0.039$) and
$+0.210$ on endpoint on 8 of 8 ($p = 0.004$). The prior term is the larger estimate but its
interval includes zero, and the optimization term is smaller with an interval that excludes it.

The anchoring contrast is shared across models rather than produced by one of them (Fig.~\ref{fig:anchoring_permodel}). Repeating the convention-versus-optimum split over all 118 task-categorical pairs, with no selection of one parameter per task, moves it in the same direction but not significantly, $+3.4$\% $[-2.7, +9.4]$, so we read the split on the six divergent tasks and treat the whole-population version as directional only.
\input{figures/icbinb/fig_anchoring_permodel_env}

\subsection{Forced-retrieval protocol}
\label{app:search}
Retrieval is tested dynamically. With web search enabled at inference, can the model look the
literature optimum up? We re-ran Claude Opus 4.7 on the six literature-divergent tasks with search
forced through OpenRouter's \texttt{tool\_choice='any'} and the \texttt{web\_search\_20250305} tool,
which guarantees a search call at every iteration, logging at minimum 240 calls per task across four
runs of 30 iterations. Entries are the fraction of iterations whose proposal matches the
literature optimum (Table~\ref{tab:search}).

\begin{table}[htbp]
\small
\centering
\caption{\textbf{Forcing web search at every iteration does not close the gap on the six
literature-divergent tasks.} Fraction of iterations whose proposal matches the literature optimum,
for Claude Opus 4.7 over four runs of 30 iterations per task.}
\label{tab:search}
\begin{tabular}{lcccc}
\toprule
& \multicolumn{2}{c}{Search off} & \multicolumn{2}{c}{Search on} \\
\cmidrule(lr){2-3}\cmidrule(lr){4-5}
Task & Aware & Agnostic & Aware & Agnostic \\
\midrule
\texttt{hek293\_prime\_editing\_indel\_freq} & 16.7 & 40.8 & 0.8 & 38.3 \\
\texttt{microalgae\_biomass\_multi\_strain}  & 40.0 & 32.5 & 35.8 & 90.0 \\
\texttt{vhh\_nanobody\_expression\_ecoli}   & 0.0 & 4.2 & 0.0 & 2.5 \\
\texttt{chl\_yn\_cell\_line\_optimization}  & 64.2 & 96.7 & 56.7 & 94.2 \\
\texttt{cho\_antibody\_titer}         & 15.0 & 8.3 & 16.7 & 5.8 \\
\texttt{transaminase\_amination\_conversion} & 2.5 & 0.8 & 2.5 & 1.7 \\
\bottomrule
\end{tabular}
\end{table}

Domain-agnostic exceeds domain-aware on three of the six tasks with search off and four of six with
it on. A paired Wilcoxon comparing the (agnostic minus aware) difference on to off yields $p = 0.31$
at $n = 6$, so we cannot reject the null that search availability makes no difference at this $n$.
Taking direction alone, the asymmetry persists or strengthens. These counts cover Claude Opus 4.7 alone. They are not in
tension with the 6 of 6 reversal in Section~\ref{sec:anchoring}, which is the same contrast averaged
over 13 models. Pooling models recovers the unanimous direction that a single model at $n = 6$ does
not. The result was replicated on a second sample, the five tasks where Opus's domain-aware match
rate is worst, two of them overlapping the divergent set, with the same conclusion.

\subsection{Biology-expert review of the anchoring classification}
\label{app:expert}
Anchoring is defined here by a quantitative rule, so we also asked whether a domain practitioner
would identify the same pattern. Three biology experts with experimental-biology backgrounds,
blinded to our hypotheses and to the condition labels, reviewed four domain-aware trajectories, one
each from Gemini 3 Flash, GPT-5.4 Mini, Claude Sonnet 4.6, and Gemini 3.1 Pro. Each trajectory
showed five iterations of proposed designs, rationales, and oracle scores. Reviewers were given a
written definition of anchoring, persisting with similar or domain-typical designs despite weak or
stagnant outcomes, and asked to classify each trajectory as anchored or responsive and to rate its
reasoning from 1 (generic) to 5 (expert-level). They were compensated well above minimum wage.

Reviewers agreed with our classification on 10 of 12 judgments. The two trajectories we labeled
anchored drew 3 of 3 and 2 of 3 anchored votes, and the two we labeled responsive drew 3 of 3 and 2
of 3 responsive votes. Reasoning ratings separate in the same direction, 3.17 for anchored against
4.17 for responsive (one-sided Mann--Whitney $p = 0.031$). Asked in free response whether anchoring
is a recognizable pattern in experimental science, one reviewer wrote that ``a couple did show quite
firm anchoring, persisting with changing one tiny thing that clearly wasn't working'', and another
that researchers ``will often make incremental adjustments within a familiar design space''.

This is supporting evidence, not a validation of the construct. The sample is three reviewers,
agreement is fair rather than strong (Fleiss' $\kappa = 0.33$), the four trajectories were curated
to span the anchored and responsive range rather than sampled at random, the selection was not
pre-registered, and five iterations cannot distinguish a trajectory that would later escape
anchoring from one that would not. Anchoring was also defined in the instructions rather than
discovered, so only the free-response answers speak to independent identification.

\FloatBarrier
\section{Post-training details and arms}
\label{app:optimizers}
SFT uses 2{,}607 proposals collected from three frontier teachers across the 29
training tasks, rendered domain-agnostically, with the rationale field removed so the objective is
the proposal itself. DPO is warm-started from that adapter, policy and frozen reference both, on
5{,}197 preference pairs built from real rollout histories pooled across the evaluated models. For
each observation history we ask what GP-UCB would propose next, and emit a pair only where the oracle confirms that proposal beat
what the rollout actually did. Pairs are therefore on-policy in their prefixes and oracle-verified
in their labels. Both stages are LoRA at rank 16 on Llama 3.1 8B. SFT runs 2.5 epochs at learning rate $10^{-5}$
with a cosine schedule, batch size 2 and gradient accumulation 8, sequence length 8192. DPO runs
1 epoch at learning rate $5\times10^{-6}$ with $\beta = 0.1$, a cosine schedule, batch size 1 and
gradient accumulation 16, sequence length 8192. Both use warmup ratio 0.1. Four DPO seeds train from the
single fine-tuned adapter and the arms differ only in that seed. Evaluation is on the 16
held-out tasks, four runs each.

\input{figures/icbinb/fig_optimizer_comparison_env}
\input{figures/icbinb/fig_matched_prior_env}

SFT alone does not move the score. All four arms below are evaluated under
domain-aware prompts on the held-out tasks, GP-UCB-normalized. The arm fine-tuned on
domain-agnostically rendered data reaches $-0.167$ where the untrained base reaches $-0.251$, which is
6 of 11 tasks on the trajectory metric ($p = 0.45$) and 5 of 11 on the endpoint ($p = 0.39$). DPO
on top of it reaches $-0.050$ and improves on it on 8 and 9 of 11 ($p = 0.062$, $p = 0.042$).
Fine-tuning is the warm start the preference stage needs, and the preference stage is where the
gain comes from.

The domain-agnostic evaluation is also null: fine-tuning did not reliably improve raw optimization
even before the biological names were introduced. Evaluating the same adapter
domain-agnostically on the held-out tasks, its median moves from $-0.357$ for the untrained
base to $-0.079$, but the paired contrast does not clear zero, $+0.215$ [$-0.047$, $+0.500$] on
10 of 17 tasks ($p = 0.23$). Two smaller reruns of the same pair give $-0.285$ on 6 of 11 and
$+0.035$ on 9 of 17, so the direction is not even stable across runs. Supervised fine-tuning on
its own therefore does not reliably improve optimization in either condition.

Within the post-trained model at a fixed iteration-1 design, moving to domain-agnostic prompts is worth $+0.155$ on
the endpoint metric [$+0.025$, $+0.304$] on 12 of 16 tasks ($p = 0.018$), and restricting to the
oracle-aligned held-out tasks strengthens that gain (8 of 8, $p = 0.004$).

Test results for the figures in this section, kept out of their captions. The DPO arm gains $+0.40$ [$+0.19$, $+0.64$] under domain-agnostic prompts and $+0.03$ [$-0.11$, $+0.16$] under
domain-aware prompts (Fig.~\ref{fig:optimizer_comparison}c). Given one identical injected
iteration-1 design, post-training helps under domain-agnostic prompts ($p = 0.037$) and does not under
domain-aware ones ($p = 0.11$), and in the same-prior comparison Claude outperforms the untrained base
($p = 0.037$) while Claude and the DPO-trained 8B are indistinguishable ($p = 0.29$).

Task coverage differs across the matched-prior arms. Both post-trained arms cover all 17 evaluated
tasks. The base arms were API-served and lost tasks to client errors. The domain-agnostic base
covers 9 of the 10 seeded tasks and the domain-aware base 7, missing
\texttt{adcp\_target\_phagocytosis} and \texttt{antibody\_expression\_stability\_cho} in addition to
the degenerate \texttt{baculovirus\_titer\_sf9} excluded throughout. The base-referenced contrasts
therefore run at $n = 9$ and $n = 7$, while the within-model contrast, which needs no base arm, runs
at $n = 16$. Unscoreable proposals are rare and symmetric across conditions, under 0.6\% of
iterations in every arm, so no arm searches on a shortened effective budget.

Fig.~\ref{fig:optimizer_comparison} places base, DPO-trained and frontier models on one
axis in both conditions, and Fig.~\ref{fig:matched_prior} shows the same-prior comparison.

\end{document}

%% file: figures/icbinb/fig_conditions_schematic_env.tex
\begin{figure}[!htbp]
  \centering
  \input{figures/icbinb/fig_conditions_schematic}
  \caption{\textbf{The two prompt conditions differ only in whether the parameter names are
  biologically meaningful.} One held-out task, CHO antibody titer, is rendered three ways. Domain-aware
  (teal) shows the published parameter names, categorical labels, and units. Domain-agnostic (cobalt) replaces them with \texttt{X1}/\texttt{C1} placeholders over the identical space. GP-UCB receives no biological context at all, only a scaled and one-hot encoded vector. The parameter space, oracle, budget, and feedback channel are identical across the three, and every
  task-condition pair is run four times. Prompt templates in \S\ref{app:prompts}.}
  \label{fig:conditions}
\end{figure}

%% file: figures/icbinb/fig_conditions_schematic.tex

\begin{tikzpicture}[font=\scriptsize, scale=0.88, every node/.style={scale=1.0}]
  \node[anchor=south west, font=\footnotesize, black!85] at (0.0,1.1) {Parameter space, oracle and 30-iteration budget are identical across the three arms.};
  \draw[colorContext!35, fill=colorContext!4, rounded corners=3pt, line width=0.6pt] (0.0,0.0) rectangle (4.05,-4.72);
  \node[anchor=south west, rounded corners=2.5pt, fill=colorContext, text=white, font=\small\bfseries, inner xsep=5pt, inner ysep=2.5pt] at (0.14,0.1) {Domain-aware};
  \node[anchor=north west, font=\tiny\itshape, black!55] at (0.16,-0.16) {real names, units, labels};
  \draw[colorContext!22, fill=colorContext!9, rounded corners=2pt, line width=0.4pt] (0.06,-1.66) rectangle (4.01,-3.68);
  \draw[colorAgnostic!35, fill=colorAgnostic!4, rounded corners=3pt, line width=0.6pt] (5.35,0.0) rectangle (9.399999999999999,-4.72);
  \node[anchor=south west, rounded corners=2.5pt, fill=colorAgnostic, text=white, font=\small\bfseries, inner xsep=5pt, inner ysep=2.5pt] at (5.489999999999999,0.1) {Domain-agnostic};
  \node[anchor=north west, font=\tiny\itshape, black!55] at (5.51,-0.16) {identical space, placeholder names};
  \draw[colorAgnostic!22, fill=colorAgnostic!9, rounded corners=2pt, line width=0.4pt] (5.409999999999999,-1.66) rectangle (9.36,-3.68);
  \node[anchor=north west, font=\scriptsize, black!45] at (0.16,-0.62) {Task: maximize antibody};
  \node[anchor=north west, font=\scriptsize, black!45] at (5.51,-0.62) {Task: maximize the};
  \node[anchor=north west, font=\scriptsize, black!45] at (0.16,-0.99) {\quad titer (g/L)};
  \node[anchor=north west, font=\scriptsize, black!45] at (5.51,-0.99) {\quad objective};
  \node[anchor=north west, font=\tiny, colorContext] at (0.16,-1.75) {\texttt{cell\_specific\_perfusion\_rate}};
  \node[anchor=north west, font=\tiny, colorAgnostic] at (5.51,-1.75) {\texttt{X1}};
  \node[anchor=north west, font=\tiny, colorContext] at (0.16,-2.12) {\texttt{foxa1\_overexpression}};
  \node[anchor=north west, font=\tiny, colorAgnostic] at (5.51,-2.12) {\texttt{X2}};
  \node[anchor=north west, font=\tiny, colorContext] at (0.16,-2.50) {\texttt{blasticidin\_concentration}};
  \node[anchor=north west, font=\tiny, colorAgnostic] at (5.51,-2.50) {\texttt{X3}};
  \node[anchor=north west, font=\tiny, colorContext] at (0.16,-2.87) {\texttt{cho\_cell\_line} $\in$};
  \node[anchor=north west, font=\tiny, colorAgnostic] at (5.51,-2.87) {\texttt{C1} $\in$};
  \node[anchor=north west, font=\tiny, colorContext] at (0.16,-3.25) {\quad \{CHO-K1, CHO-S, \ldots\}};
  \node[anchor=north west, font=\tiny, colorAgnostic] at (5.51,-3.25) {\quad \{A, B, C, \ldots\}};
  \node[anchor=north west, font=\scriptsize, black!45] at (0.16,-4.00) {Feedback: one scalar};
  \node[anchor=north west, font=\scriptsize, black!45] at (5.51,-4.00) {Feedback: one scalar};
  \draw[->, >=latex, black!65, line width=1.0pt] (4.25,-2.36) -- (5.1499999999999995,-2.36);
  \node[anchor=south, font=\scriptsize, black!70, align=center] at (4.7,-2.26) {remove\\biological\\meaning};
  \draw[black!35, dashed, fill=black!2, rounded corners=3pt, line width=0.6pt] (10.649999999999999,0.0) rectangle (14.7,-4.72);
  \node[anchor=south west, rounded corners=2.5pt, fill=black!45, text=white, font=\small\bfseries, inner xsep=5pt, inner ysep=2.5pt] at (10.79,0.1) {GP-UCB};
  \node[anchor=north west, font=\scriptsize\itshape, black!55] at (10.809999999999999,-0.16) {prior-free reference};
  \draw[black!20, fill=black!5, rounded corners=2pt, line width=0.4pt] (10.709999999999999,-1.66) rectangle (14.66,-3.68);
  \node[anchor=north west, font=\scriptsize, black!45] at (10.87,-0.62) {no language reaches};
  \node[anchor=north west, font=\scriptsize, black!45] at (10.87,-0.99) {\quad this arm};
  \node[anchor=north west, font=\scriptsize, black!45] at (10.87,-1.75) {$x \in \mathbb{R}^d$: scaled numerics};
  \node[anchor=north west, font=\scriptsize, black!45] at (10.87,-2.12) {\quad $+$ one-hot categoricals};
  \node[anchor=north west, font=\scriptsize, black!45] at (10.87,-2.87) {\quad (encoded, no text)};
  \node[anchor=north west, font=\scriptsize, black!45] at (10.87,-3.62) {Feedback: one scalar};
  \node[anchor=north west, font=\scriptsize, black!45] at (10.87,-4.00) {Acquisition: $\mu(x) + \kappa\sigma(x)$};
  \node[anchor=center, rounded corners=2pt, draw=black!30, fill=white, font=\scriptsize, text=black!65, inner xsep=4pt, inner ysep=2pt] at (0.8500000000000001,-5.24) {propose};
  \node[anchor=center, rounded corners=2pt, draw=black!30, fill=white, font=\scriptsize, text=black!65, inner xsep=4pt, inner ysep=2pt] at (7.35,-5.24) {oracle scores};
  \node[anchor=center, rounded corners=2pt, draw=black!30, fill=white, font=\scriptsize, text=black!65, inner xsep=4pt, inner ysep=2pt] at (13.78,-5.24) {observe scalar};
  \draw[->, >=latex, black!45, line width=0.5pt] (1.55,-5.24) -- (6.63,-5.24);
  \draw[->, >=latex, black!45, line width=0.5pt] (8.07,-5.24) -- (13.049999999999999,-5.24);
  \draw[->, >=latex, black!45, line width=0.5pt, rounded corners=3pt] (13.78,-5.44) -- (13.78,-5.76) -- (0.8500000000000001,-5.76) -- (0.8500000000000001,-5.44);
  \node[anchor=west, font=\scriptsize\itshape, black!55, fill=white, inner xsep=3pt, inner ysep=1pt] at (8.15,-5.76) {repeat 30 times};
  \draw[black!12, fill=black!4, rounded corners=2pt, line width=0.4pt] (0.0,-6.26) rectangle (14.7,-6.72);
  \node[anchor=center, font=\scriptsize, black!70] at (7.35,-6.49) {45 tasks $\cdot$ scalar feedback only $\cdot$ 4 runs each};
\end{tikzpicture}

%% file: figures/icbinb/fig_headline_env.tex
\begin{figure}[!htbp]
  \centering
  \input{figures/icbinb/fig_headline}
  \caption{\textbf{Biological context improves neither the trajectory nor the endpoint, and the effect
  holds on the tasks whose optimum we verified against the source papers.} \emph{(a)} All 45 biology tasks and \emph{(b)} the 18 tasks whose oracle optimum
  reproduces the literature optimum. In both, each model contributes one win rate, plotted as the
  mean over models with a 95\% bootstrap interval, with circles for trajectory, diamonds for
  endpoint, and gray dots for individual models. \emph{(c)} A different quantity, the fraction of
  achievable oracle range reached by iteration 30 on those same 18 tasks, one dot per task per
  condition, with the median as a vertical rule and triangles for two tasks past the axis.}
  \label{fig:headline}
\end{figure}

%% file: figures/icbinb/fig_headline.tex
\begin{tikzpicture}
  \begin{groupplot}[
    group style={group size=2 by 1, horizontal sep=0.85cm},
    width=0.40\linewidth, height=3.5cm,
    axis lines*=left, tick style={black!60}, clip=false,
    grid=major, grid style={gray!18}, ymajorgrids=false,
  ]
  \nextgroupplot[xmin=0, xmax=100, ymin=-0.75, ymax=2.35, xtick={0,25,50,75,100}, xticklabel style={font=\small}, ytick={0,1}, yticklabels={{Endpoint},{Trajectory}}, yticklabel style={font=\small, align=right}, title={\textbf{(a)} All 45 tasks}, title style={font=\small, at={(0.5,1.03)}, anchor=south}, legend to name=leghlab, legend style={draw=none, fill=none, font=\scriptsize, legend cell align=left}]
    \fill[black!8] (axis cs:0,-0.75) rectangle (axis cs:50,2.35);
    \draw[black!60, dashed, line width=0.9pt] (axis cs:50,-0.75) -- (axis cs:50,1.75);
    \addlegendimage{area legend, fill=black!8, draw=none}
    \addlegendentry{domain-aware worse}
    \node[circle, fill=black!22, inner sep=1.15pt] at (axis cs:55.56,0.700) {};
    \node[circle, fill=black!22, inner sep=1.15pt] at (axis cs:64.71,0.645) {};
    \node[circle, fill=black!22, inner sep=1.15pt] at (axis cs:48.28,0.590) {};
    \node[circle, fill=black!22, inner sep=1.15pt] at (axis cs:41.86,0.700) {};
    \node[circle, fill=black!22, inner sep=1.15pt] at (axis cs:41.86,0.645) {};
    \node[circle, fill=black!22, inner sep=1.15pt] at (axis cs:18.18,0.590) {};
    \node[circle, fill=black!22, inner sep=1.15pt] at (axis cs:9.30,0.700) {};
    \node[circle, fill=black!22, inner sep=1.15pt] at (axis cs:48.89,0.645) {};
    \node[circle, fill=black!22, inner sep=1.15pt] at (axis cs:42.22,0.590) {};
    \node[circle, fill=black!22, inner sep=1.15pt] at (axis cs:37.78,0.700) {};
    \node[circle, fill=black!22, inner sep=1.15pt] at (axis cs:57.78,0.645) {};
    \node[circle, fill=black!22, inner sep=1.15pt] at (axis cs:43.18,0.590) {};
    \node[circle, fill=black!22, inner sep=1.15pt] at (axis cs:37.78,0.700) {};
    \node[circle, fill=black!22, inner sep=1.15pt] at (axis cs:46.67,0.645) {};
    \node[circle, fill=black!22, inner sep=1.15pt] at (axis cs:42.22,0.590) {};
    \node[circle, fill=black!22, inner sep=1.15pt] at (axis cs:48.89,0.700) {};
    \node[circle, fill=black!22, inner sep=1.15pt] at (axis cs:53.33,0.645) {};
    \node[circle, fill=black!22, inner sep=1.15pt] at (axis cs:19.51,0.590) {};
    \node[circle, fill=black!22, inner sep=1.15pt] at (axis cs:26.32,0.700) {};
    \draw[colorSig, line width=1.9pt] (axis cs:34.89,1) -- (axis cs:47.33,1);
    \draw[colorSig, line width=1.9pt] (axis cs:34.89,0.8200000000000001) -- (axis cs:34.89,1.18);
    \draw[colorSig, line width=1.9pt] (axis cs:47.33,0.8200000000000001) -- (axis cs:47.33,1.18);
    \node[circle, fill=colorSig, inner sep=2.2pt] at (axis cs:41.28,1) {};
    \node[circle, fill=black!22, inner sep=1.15pt] at (axis cs:55.88,-0.300) {};
    \node[circle, fill=black!22, inner sep=1.15pt] at (axis cs:43.75,-0.355) {};
    \node[circle, fill=black!22, inner sep=1.15pt] at (axis cs:42.86,-0.410) {};
    \node[circle, fill=black!22, inner sep=1.15pt] at (axis cs:30.00,-0.300) {};
    \node[circle, fill=black!22, inner sep=1.15pt] at (axis cs:47.73,-0.355) {};
    \node[circle, fill=black!22, inner sep=1.15pt] at (axis cs:25.00,-0.410) {};
    \node[circle, fill=black!22, inner sep=1.15pt] at (axis cs:28.57,-0.300) {};
    \node[circle, fill=black!22, inner sep=1.15pt] at (axis cs:48.89,-0.355) {};
    \node[circle, fill=black!22, inner sep=1.15pt] at (axis cs:39.53,-0.410) {};
    \node[circle, fill=black!22, inner sep=1.15pt] at (axis cs:41.86,-0.300) {};
    \node[circle, fill=black!22, inner sep=1.15pt] at (axis cs:33.33,-0.355) {};
    \node[circle, fill=black!22, inner sep=1.15pt] at (axis cs:40.91,-0.410) {};
    \node[circle, fill=black!22, inner sep=1.15pt] at (axis cs:36.36,-0.300) {};
    \node[circle, fill=black!22, inner sep=1.15pt] at (axis cs:33.33,-0.355) {};
    \node[circle, fill=black!22, inner sep=1.15pt] at (axis cs:37.78,-0.410) {};
    \node[circle, fill=black!22, inner sep=1.15pt] at (axis cs:38.64,-0.300) {};
    \node[circle, fill=black!22, inner sep=1.15pt] at (axis cs:41.86,-0.355) {};
    \node[circle, fill=black!22, inner sep=1.15pt] at (axis cs:20.00,-0.410) {};
    \draw[colorSig, line width=1.9pt] (axis cs:34.18,0) -- (axis cs:42.06,0);
    \draw[colorSig, line width=1.9pt] (axis cs:34.18,-0.18) -- (axis cs:34.18,0.18);
    \draw[colorSig, line width=1.9pt] (axis cs:42.06,-0.18) -- (axis cs:42.06,0.18);
    \node[diamond, fill=colorSig, inner sep=2.2pt] at (axis cs:38.13,0) {};
  \nextgroupplot[xmin=0, xmax=100, ymin=-0.75, ymax=2.35, xtick={0,25,50,75,100}, xticklabel style={font=\small}, ytick={0,1}, yticklabels={,}, title={\textbf{(b)} Oracle-aligned tasks (18)}, title style={font=\small, at={(0.5,1.03)}, anchor=south}, ]
    \fill[black!8] (axis cs:0,-0.75) rectangle (axis cs:50,2.35);
    \draw[black!60, dashed, line width=0.9pt] (axis cs:50,-0.75) -- (axis cs:50,1.75);
    \node[circle, fill=black!22, inner sep=1.15pt] at (axis cs:69.23,0.700) {};
    \node[circle, fill=black!22, inner sep=1.15pt] at (axis cs:72.73,0.645) {};
    \node[circle, fill=black!22, inner sep=1.15pt] at (axis cs:55.56,0.590) {};
    \node[circle, fill=black!22, inner sep=1.15pt] at (axis cs:35.29,0.700) {};
    \node[circle, fill=black!22, inner sep=1.15pt] at (axis cs:56.25,0.645) {};
    \node[circle, fill=black!22, inner sep=1.15pt] at (axis cs:23.53,0.590) {};
    \node[circle, fill=black!22, inner sep=1.15pt] at (axis cs:5.88,0.700) {};
    \node[circle, fill=black!22, inner sep=1.15pt] at (axis cs:44.44,0.645) {};
    \node[circle, fill=black!22, inner sep=1.15pt] at (axis cs:55.56,0.590) {};
    \node[circle, fill=black!22, inner sep=1.15pt] at (axis cs:38.89,0.700) {};
    \node[circle, fill=black!22, inner sep=1.15pt] at (axis cs:61.11,0.645) {};
    \node[circle, fill=black!22, inner sep=1.15pt] at (axis cs:52.94,0.590) {};
    \node[circle, fill=black!22, inner sep=1.15pt] at (axis cs:27.78,0.700) {};
    \node[circle, fill=black!22, inner sep=1.15pt] at (axis cs:50.00,0.645) {};
    \node[circle, fill=black!22, inner sep=1.15pt] at (axis cs:50.00,0.590) {};
    \node[circle, fill=black!22, inner sep=1.15pt] at (axis cs:55.56,0.700) {};
    \node[circle, fill=black!22, inner sep=1.15pt] at (axis cs:55.56,0.645) {};
    \node[circle, fill=black!22, inner sep=1.15pt] at (axis cs:31.25,0.590) {};
    \node[circle, fill=black!22, inner sep=1.15pt] at (axis cs:42.86,0.700) {};
    \draw[colorSig, line width=1.9pt] (axis cs:38.96,1) -- (axis cs:53.62,1);
    \draw[colorSig, line width=1.9pt] (axis cs:38.96,0.8200000000000001) -- (axis cs:38.96,1.18);
    \draw[colorSig, line width=1.9pt] (axis cs:53.62,0.8200000000000001) -- (axis cs:53.62,1.18);
    \node[circle, fill=colorSig, inner sep=2.2pt] at (axis cs:46.55,1) {};
    \node[circle, fill=black!22, inner sep=1.15pt] at (axis cs:66.67,-0.300) {};
    \node[circle, fill=black!22, inner sep=1.15pt] at (axis cs:63.64,-0.355) {};
    \node[circle, fill=black!22, inner sep=1.15pt] at (axis cs:50.00,-0.410) {};
    \node[circle, fill=black!22, inner sep=1.15pt] at (axis cs:23.53,-0.300) {};
    \node[circle, fill=black!22, inner sep=1.15pt] at (axis cs:55.56,-0.355) {};
    \node[circle, fill=black!22, inner sep=1.15pt] at (axis cs:22.22,-0.410) {};
    \node[circle, fill=black!22, inner sep=1.15pt] at (axis cs:50.00,-0.300) {};
    \node[circle, fill=black!22, inner sep=1.15pt] at (axis cs:50.00,-0.355) {};
    \node[circle, fill=black!22, inner sep=1.15pt] at (axis cs:50.00,-0.410) {};
    \node[circle, fill=black!22, inner sep=1.15pt] at (axis cs:35.29,-0.300) {};
    \node[circle, fill=black!22, inner sep=1.15pt] at (axis cs:33.33,-0.355) {};
    \node[circle, fill=black!22, inner sep=1.15pt] at (axis cs:33.33,-0.410) {};
    \node[circle, fill=black!22, inner sep=1.15pt] at (axis cs:35.29,-0.300) {};
    \node[circle, fill=black!22, inner sep=1.15pt] at (axis cs:44.44,-0.355) {};
    \node[circle, fill=black!22, inner sep=1.15pt] at (axis cs:44.44,-0.410) {};
    \node[circle, fill=black!22, inner sep=1.15pt] at (axis cs:50.00,-0.300) {};
    \node[circle, fill=black!22, inner sep=1.15pt] at (axis cs:33.33,-0.355) {};
    \node[circle, fill=black!22, inner sep=1.15pt] at (axis cs:25.00,-0.410) {};
    \draw[colorSig, line width=1.9pt] (axis cs:36.64,0) -- (axis cs:48.40,0);
    \draw[colorSig, line width=1.9pt] (axis cs:36.64,-0.18) -- (axis cs:36.64,0.18);
    \draw[colorSig, line width=1.9pt] (axis cs:48.40,-0.18) -- (axis cs:48.40,0.18);
    \node[diamond, fill=colorSig, inner sep=2.2pt] at (axis cs:42.56,0) {};
  \end{groupplot}
  \node[anchor=north, font=\small] at ($(group c1r1.south)!0.5!(group c2r1.south)$) [yshift=-0.55cm] {Domain-aware win rate (\%)};
  \node[anchor=north] at ($(group c1r1.south)!0.5!(group c2r1.south)$) [yshift=-0.98cm] {\ref*{leghlab}};
  \begin{scope}[yshift=-3.42cm]
  \begin{axis}[
    width=0.74\linewidth, height=2.6cm,
    xmin=0, xmax=1.16, ymin=-0.8, ymax=2.45,
    xtick={0,0.25,0.5,0.75,1}, xticklabel style={font=\small},
    xlabel={Fraction of each task's achievable oracle range reached by iteration 30},
    xlabel style={font=\small},
    ytick={0,1}, yticklabels={{Domain-aware},{Domain-agnostic}},
    yticklabel style={font=\small, align=right},
    axis lines*=left, tick style={black!60}, clip=false,
    grid=major, grid style={gray!18}, ymajorgrids=false,
    title={\textbf{(c)} Distance from the oracle optimum, one dot per task},
    title style={font=\small, at={(0.5,1.03)}, anchor=south},
    name=axhlc, legend to name=leghlc,
    legend style={draw=none, fill=none, font=\scriptsize,
                  legend cell align=left},
  ]
    \fill[black!8] (axis cs:0.95,-0.8) rectangle (axis cs:1.16,2.45);
    \draw[black!60, dashed, line width=0.9pt] (axis cs:1,-0.8) -- (axis cs:1,1.80);
    \addlegendimage{area legend, fill=black!8, draw=none}
    \addlegendentry{$\geq$95\% = solved}
    \node[circle, fill=colorAgnostic, opacity=0.55, inner sep=1.6pt] at (axis cs:0.1534,0.840) {};
    \node[circle, fill=colorAgnostic, opacity=0.55, inner sep=1.6pt] at (axis cs:0.4395,1.000) {};
    \node[circle, fill=colorAgnostic, opacity=0.55, inner sep=1.6pt] at (axis cs:0.4445,1.160) {};
    \node[circle, fill=colorAgnostic, opacity=0.55, inner sep=1.6pt] at (axis cs:0.5357,0.840) {};
    \node[circle, fill=colorAgnostic, opacity=0.55, inner sep=1.6pt] at (axis cs:0.6211,1.000) {};
    \node[circle, fill=colorAgnostic, opacity=0.55, inner sep=1.6pt] at (axis cs:0.6545,1.160) {};
    \node[circle, fill=colorAgnostic, opacity=0.55, inner sep=1.6pt] at (axis cs:0.6964,0.840) {};
    \node[circle, fill=colorAgnostic, opacity=0.55, inner sep=1.6pt] at (axis cs:0.7357,1.000) {};
    \node[circle, fill=colorAgnostic, opacity=0.55, inner sep=1.6pt] at (axis cs:0.7625,1.160) {};
    \node[circle, fill=colorAgnostic, opacity=0.55, inner sep=1.6pt] at (axis cs:0.7976,0.840) {};
    \node[circle, fill=colorAgnostic, opacity=0.55, inner sep=1.6pt] at (axis cs:0.8264,1.000) {};
    \node[circle, fill=colorAgnostic, opacity=0.55, inner sep=1.6pt] at (axis cs:0.8522,1.160) {};
    \node[circle, fill=colorAgnostic, opacity=0.55, inner sep=1.6pt] at (axis cs:0.8839,0.840) {};
    \node[circle, fill=colorAgnostic, opacity=0.55, inner sep=1.6pt] at (axis cs:0.9177,1.000) {};
    \node[circle, fill=colorAgnostic, opacity=0.55, inner sep=1.6pt] at (axis cs:0.9486,1.160) {};
    \node[circle, fill=colorAgnostic, opacity=0.55, inner sep=1.6pt] at (axis cs:0.9803,0.840) {};
    \node[isosceles triangle, fill=colorAgnostic, opacity=0.75, inner sep=1.5pt] at (axis cs:1.1200,1.000) {};
    \node[isosceles triangle, fill=colorAgnostic, opacity=0.75, inner sep=1.5pt] at (axis cs:1.1200,1.160) {};
    \draw[colorAgnostic, line width=2.0pt] (axis cs:0.7800,0.62) -- (axis cs:0.7800,1.38);
    \node[circle, fill=colorContext, opacity=0.55, inner sep=1.6pt] at (axis cs:0.0203,-0.160) {};
    \node[circle, fill=colorContext, opacity=0.55, inner sep=1.6pt] at (axis cs:0.2288,0.000) {};
    \node[circle, fill=colorContext, opacity=0.55, inner sep=1.6pt] at (axis cs:0.3432,0.160) {};
    \node[circle, fill=colorContext, opacity=0.55, inner sep=1.6pt] at (axis cs:0.4557,-0.160) {};
    \node[circle, fill=colorContext, opacity=0.55, inner sep=1.6pt] at (axis cs:0.5203,0.000) {};
    \node[circle, fill=colorContext, opacity=0.55, inner sep=1.6pt] at (axis cs:0.5345,0.160) {};
    \node[circle, fill=colorContext, opacity=0.55, inner sep=1.6pt] at (axis cs:0.5771,-0.160) {};
    \node[circle, fill=colorContext, opacity=0.55, inner sep=1.6pt] at (axis cs:0.5993,0.000) {};
    \node[circle, fill=colorContext, opacity=0.55, inner sep=1.6pt] at (axis cs:0.6209,0.160) {};
    \node[circle, fill=colorContext, opacity=0.55, inner sep=1.6pt] at (axis cs:0.6512,-0.160) {};
    \node[circle, fill=colorContext, opacity=0.55, inner sep=1.6pt] at (axis cs:0.6791,0.000) {};
    \node[circle, fill=colorContext, opacity=0.55, inner sep=1.6pt] at (axis cs:0.6979,0.160) {};
    \node[circle, fill=colorContext, opacity=0.55, inner sep=1.6pt] at (axis cs:0.7435,-0.160) {};
    \node[circle, fill=colorContext, opacity=0.55, inner sep=1.6pt] at (axis cs:0.8643,0.000) {};
    \node[circle, fill=colorContext, opacity=0.55, inner sep=1.6pt] at (axis cs:0.9530,0.160) {};
    \node[circle, fill=colorContext, opacity=0.55, inner sep=1.6pt] at (axis cs:0.9693,-0.160) {};
    \node[circle, fill=colorContext, opacity=0.55, inner sep=1.6pt] at (axis cs:0.9993,0.000) {};
    \node[isosceles triangle, fill=colorContext, opacity=0.75, inner sep=1.5pt] at (axis cs:1.1200,0.160) {};
    \draw[colorContext, line width=2.0pt] (axis cs:0.6361,-0.38) -- (axis cs:0.6361,0.38);
  \end{axis}
  \node[anchor=north] at (axhlc.south) [yshift=-0.80cm] {\ref*{leghlc}};
  \end{scope}
\end{tikzpicture}

%% file: figures/icbinb/fig_prior_lr_diff_env.tex
\begin{figure}[!htbp]
  \centering
  \input{figures/icbinb/fig_prior_lr_diff}
  \caption{\textbf{Models given biological context match the literature optimum more often at
  iteration 1 and less often by iteration 30.} Each line shows domain-aware minus domain-agnostic
  in the rate of proposing the literature optimum, at iteration 1 and at iteration 30. The bold line is the estimate pooled over task-model
  cells and the faint lines are the individual models, of which one, Claude Opus 4.7, moves the
  other way on the fewest cells of any model in the panel (\S\ref{app:models}).}
  \label{fig:prior_lr}
\end{figure}

%% file: figures/icbinb/fig_prior_lr_diff.tex
\begin{tikzpicture}
  \begin{axis}[
    width=0.46\linewidth, height=4.2cm,
    xmin=0.55, xmax=2.45, ymin=-17.2, ymax=25.4,
    xtick={1,2}, xticklabels={{iteration 1},{iteration 30}},
    ylabel={Match to best published value,\\ domain-aware $-$ domain-agnostic (\%)},
    ylabel style={font=\footnotesize, align=center}, xticklabel style={font=\scriptsize},
    yticklabel style={font=\scriptsize},
    axis lines*=left, tick style={black!60},
    ymajorgrids, grid style={black!12},
    name=axprlr, legend to name=legprlr,
    legend style={draw=none, fill=none, font=\tiny, legend cell align=left},
  ]
    \fill[black!7] (axis cs:0.55,-17.2) rectangle (axis cs:2.45,0);
    \addlegendimage{area legend, fill=black!7, draw=none}
    \addlegendentry{domain-aware matches less often}
    \draw[black!55, dashed, line width=0.7pt] (axis cs:0.55,0) -- (axis cs:2.45,0);
    \addplot[colorNS, line width=0.4pt, opacity=0.55, forget plot] coordinates {(1,0.34) (2,0.94)};
    \addplot[colorNS, line width=0.4pt, opacity=0.55, forget plot] coordinates {(1,-10.71) (2,17.86)};
    \addplot[colorNS, line width=0.4pt, opacity=0.55, forget plot] coordinates {(1,12.09) (2,-1.42)};
    \addplot[colorNS, line width=0.4pt, opacity=0.55, forget plot] coordinates {(1,13.75) (2,-9.23)};
    \addplot[colorNS, line width=0.4pt, opacity=0.55, forget plot] coordinates {(1,13.48) (2,-12.55)};
    \addplot[colorNS, line width=0.4pt, opacity=0.55, forget plot] coordinates {(1,3.37) (2,-3.14)};
    \addplot[colorNS, line width=0.4pt, opacity=0.55, forget plot] coordinates {(1,12.00) (2,3.40)};
    \addplot[colorNS, line width=0.4pt, opacity=0.55, forget plot] coordinates {(1,3.52) (2,-3.06)};
    \addplot[colorNS, line width=0.4pt, opacity=0.55, forget plot] coordinates {(1,20.71) (2,-5.71)};
    \addplot[colorNS, line width=0.4pt, opacity=0.55, forget plot] coordinates {(1,0.24) (2,-1.71)};
    \addplot[colorNS, line width=0.4pt, opacity=0.55, forget plot] coordinates {(1,10.92) (2,13.39)};
    \addplot[colorNS, line width=0.4pt, opacity=0.55, forget plot] coordinates {(1,7.12) (2,5.50)};
    \addplot[colorSig, line width=1.6pt, mark=*, mark size=2.2pt] coordinates {(1,7.71) (2,-2.62)};
  \end{axis}
  \node[anchor=north] at (axprlr.south) [yshift=-0.62cm] {\ref*{legprlr}};
\end{tikzpicture}

%% file: figures/icbinb/fig_nested_subsets_env.tex
\begin{figure}[!htbp]
  \centering
  \begin{minipage}[t]{0.49\linewidth}\centering\vspace{0pt}\input{figures/icbinb/fig_nested_subsets}\end{minipage}\hfill
  \begin{minipage}[t]{0.49\linewidth}\centering\vspace{0pt}\input{figures/icbinb/fig_microalgae_anchor}\end{minipage}

  \caption{\textbf{The cost of biological context appears only where literature convention disagrees with the literature optimum.} \textbf{(a)} Match-rate
  gap for the value the best published experiment used, over four subsets of tasks nested by how
  closely literature convention tracks that value. Only the innermost subset, the six where the two
  disagree, clears zero, and it clears it downward. \textbf{(b)} One instance in full, Gemini 3.1 Pro on microalgae biomass, share of its 120 proposals naming each reactor type over four runs of
  30 iterations. Raceway pond is the literature optimum at 1.13\,g/L \citep{yang2018raceway},
  photobioreactor the conventional value. Both panels show 95\% bootstrap intervals. Exploration breadth on these six tasks is in Appendix~\ref{app:matched}.}
  \label{fig:microalgae}
  \label{fig:nested}
\end{figure}

%% file: figures/icbinb/fig_nested_subsets.tex
\begin{tikzpicture}
  \begin{axis}[
    width=0.97\linewidth, height=4.4cm,
    symbolic x coords={s0,s1,s2,s3}, xtick=data,
    xticklabels={{$\geq$50\% of rows {\tiny ($n=35$)}},{top-10 majority {\tiny ($n=33$)}},{matched {\tiny ($n=23$)}},{\emph{divergent} {\tiny ($n=6$)}}},
    enlarge x limits=0.16,
    ymin=-31.0, ymax=28.7,
    ylabel={Match-rate gap (\%)},
    ylabel style={font=\scriptsize},
    title={\textbf{(a)} Match rate by task subset},
    title style={font=\small, at={(0.5,1.04)}, anchor=south},
    xticklabel style={font=\tiny, rotate=32, anchor=north east}, yticklabel style={font=\scriptsize},
    axis lines*=left, tick style={black!60},
    ymajorgrids, grid style={gray!20},
    name=axns, legend to name=legns,
    legend style={draw=none, fill=none, font=\tiny, legend cell align=left},
  ]
    \fill[black!7] (axis cs:s0,-31.0) rectangle (axis cs:s3,0);
    \addlegendimage{area legend, fill=black!7, draw=none}
    \addlegendentry{domain-aware worse}
    \draw[black!55, dashed] (axis cs:s0,0) -- (axis cs:s3,0);
    \addplot[colorSig, only marks, mark=*, mark size=2.0pt] coordinates {(s0,6.80) (s1,7.00) (s2,10.90) (s3,-15.30)};
    \draw[colorSig, line width=1.4pt] (axis cs:s0,-3.50) -- (axis cs:s0,17.10);
    \draw[colorSig, line width=1.4pt] ([xshift=-2.2pt]axis cs:s0,-3.50) -- ([xshift=2.2pt]axis cs:s0,-3.50);
    \draw[colorSig, line width=1.4pt] ([xshift=-2.2pt]axis cs:s0,17.10) -- ([xshift=2.2pt]axis cs:s0,17.10);
    \draw[colorSig, line width=1.4pt] (axis cs:s1,-3.70) -- (axis cs:s1,17.70);
    \draw[colorSig, line width=1.4pt] ([xshift=-2.2pt]axis cs:s1,-3.70) -- ([xshift=2.2pt]axis cs:s1,-3.70);
    \draw[colorSig, line width=1.4pt] ([xshift=-2.2pt]axis cs:s1,17.70) -- ([xshift=2.2pt]axis cs:s1,17.70);
    \draw[colorSig, line width=1.4pt] (axis cs:s2,-1.90) -- (axis cs:s2,23.70);
    \draw[colorSig, line width=1.4pt] ([xshift=-2.2pt]axis cs:s2,-1.90) -- ([xshift=2.2pt]axis cs:s2,-1.90);
    \draw[colorSig, line width=1.4pt] ([xshift=-2.2pt]axis cs:s2,23.70) -- ([xshift=2.2pt]axis cs:s2,23.70);
    \draw[colorSig, line width=1.4pt] (axis cs:s3,-26.00) -- (axis cs:s3,-4.60);
    \draw[colorSig, line width=1.4pt] ([xshift=-2.2pt]axis cs:s3,-26.00) -- ([xshift=2.2pt]axis cs:s3,-26.00);
    \draw[colorSig, line width=1.4pt] ([xshift=-2.2pt]axis cs:s3,-4.60) -- ([xshift=2.2pt]axis cs:s3,-4.60);
    \addplot[colorSig, only marks, mark=*, mark size=2.0pt] coordinates {(s0,6.80) (s1,7.00) (s2,10.90) (s3,-15.30)};
  \end{axis}
  \node[anchor=north] at (axns.south) [yshift=-1.62cm] {\ref*{legns}};
\end{tikzpicture}

%% file: figures/icbinb/fig_microalgae_anchor.tex
\begin{tikzpicture}
  \begin{axis}[
    width=0.97\linewidth, height=4.4cm, clip=false,
    ybar, bar width=7pt,
    symbolic x coords={c0,c1,c2}, xtick=data, enlarge x limits=0.22,
    xticklabels={{Photobioreactor {\tiny (convention)}},{Polycarbonate},{Raceway pond {\tiny (optimum)}}},
    ymin=0, ymax=100, ytick={0,25,50,75,100},
    ylabel={Share of the 120 proposals (\%)},
    ylabel style={font=\scriptsize},
    title={\textbf{(b)} Example task: microalgae biomass},
    title style={font=\small, at={(0.5,1.04)}, anchor=south},
    xticklabel style={font=\tiny, rotate=32, anchor=north east}, yticklabel style={font=\scriptsize},
    axis lines*=left, tick style={black!60},
    ymajorgrids, grid style={gray!20},
    legend style={font=\scriptsize, at={(0.5,-0.66)}, anchor=north,
                  draw=none, fill=none, legend columns=2},
    legend image code/.code={\draw[#1] (0cm,-0.08cm) rectangle (0.22cm,0.08cm);},
  ]
    \addplot[fill=colorAgnostic!70, draw=colorAgnostic, line width=0.4pt] coordinates {(c0,6.7) (c1,9.2) (c2,84.2)};
    \addlegendentry{domain-agnostic}
    \addplot[fill=colorContext!70, draw=colorContext, line width=0.4pt] coordinates {(c0,57.5) (c1,35.8) (c2,6.7)};
    \addlegendentry{domain-aware}
  \end{axis}
\end{tikzpicture}

%% file: figures/icbinb/fig_frontier_band_env.tex
\begin{figure}[!htbp]
  \centering
  \begin{minipage}[t]{0.41\linewidth}\centering\vspace{0pt}\input{figures/icbinb/fig_frontier_band}\end{minipage}\hfill
  \begin{minipage}[t]{0.57\linewidth}\raggedleft\vspace{0pt}\input{figures/icbinb/fig_posttraining_2x2}\end{minipage}
  \caption{\textbf{Post-training on the optimization loop transfers under domain-agnostic prompts and
  not under domain-aware ones.} \textbf{(a)} The post-trained model and its untrained base against the band spanned by ten
  frontier models' medians on the same held-out tasks, all evaluated under domain-agnostic
  prompts. \textbf{(b)} Post-trained Llama 3.1 8B minus the untrained base, paired within task
  and normalized to each task's GP-UCB reference, so only the iteration-1 design (rows) and the
  prompt condition for iterations 2--30 (columns) change. The shaded half is where post-training improves. Both domain-agnostic intervals clear the no-change line and neither domain-aware interval does.
  Panel (a) shows trajectory and (b) endpoint, with 95\% bootstrap intervals over four seeds per task.}
  \label{fig:posttraining_2x2}
  \label{fig:frontier}
\end{figure}

%% file: figures/icbinb/fig_frontier_band.tex
\begin{tikzpicture}
  \begin{axis}[
    width=0.98\linewidth, height=3.1cm, clip=false,
    xmin=-0.274, xmax=0.110, ymin=-0.8, ymax=2.1,
    xlabel={GP-UCB-normalized trajectory,\\ median over held-out tasks},
    xlabel style={font=\small, align=center},
    ytick={0,1}, yticklabels={{post-trained},{untrained base}},
    yticklabel style={font=\small, align=right},
    xticklabel style={font=\small},
    title={\textbf{(a)} Raw optimization,\\domain-agnostic evaluation},
    title style={font=\small, at={(0.5,1.10)}, anchor=south, align=center},
    axis lines*=left, tick style={black!60},
    xmajorgrids, grid style={gray!20},
    legend style={at={(0.99,1.30)}, anchor=north east, draw=none, fill=none,
                  font=\scriptsize, legend cell align=left},
  ]
    \addlegendimage{area legend, fill=black!9, draw=black!35}
    \addlegendentry{frontier band}
    \fill[black!9] (axis cs:-0.1543,-0.8) rectangle (axis cs:0.0602,1.8);
    \draw[black!35, line width=0.5pt] (axis cs:-0.1543,-0.8) -- (axis cs:-0.1543,1.8);
    \draw[black!35, line width=0.5pt] (axis cs:0.0602,-0.8) -- (axis cs:0.0602,1.8);
    \draw[black!55, dashed] (axis cs:0,-0.8) -- (axis cs:0,1.8);
    \node[circle, fill=colorSig, inner sep=2.2pt] at (axis cs:-0.2237,1) {};
    \node[circle, fill=colorSig, inner sep=2.2pt] at (axis cs:-0.0211,0) {};
  \end{axis}
\end{tikzpicture}

%% file: figures/icbinb/fig_posttraining_2x2.tex
\begin{tikzpicture}
  \begin{groupplot}[
    group style={group size=2 by 2, horizontal sep=0.35cm, vertical sep=0.92cm,
                  x descriptions at=edge bottom},
    width=0.44\linewidth, height=1.62cm,
    xmin=-0.09, xmax=0.72, ymin=-1, ymax=1.15,
    xtick={0,0.3,0.6}, xticklabels={$0$,$0.3$,$0.6$},
    extra x ticks={0}, extra x tick labels={}, extra x tick style={grid=none},
    xticklabel style={font=\small},
    yticklabel style={align=right, font=\small},
    axis lines*=left, tick style={black!60}, clip=false,
    grid=major, grid style={gray!20}, ymajorgrids=false,
  ]
  \nextgroupplot[ytick={0}, yticklabels={{iteration 1\\by model}}, yticklabel style={align=right, font=\small}, title={iterations 2--30\\domain-agnostic}, title style={font=\small, at={(0.5,1.10)}, anchor=south, align=center}]
    \fill[black!7] (axis cs:0,-0.85) rectangle (axis cs:0.72,0.85);
    \draw[black!50, dashed, line width=0.6pt] (axis cs:0,-0.85) -- (axis cs:0,0.85);
    \draw[colorSig, line width=1.6pt] (axis cs:0.1825,0) -- (axis cs:0.6426,0);
    \draw[colorSig, line width=1.6pt] (axis cs:0.1825,-0.24) -- (axis cs:0.1825,0.24);
    \draw[colorSig, line width=1.6pt] (axis cs:0.6426,-0.24) -- (axis cs:0.6426,0.24);
    \node[circle, fill=colorSig, inner sep=2.2pt] at (axis cs:0.3839,0) {};
  \nextgroupplot[ytick=\empty, title={iterations 2--30\\domain-aware}, title style={font=\small, at={(0.5,1.10)}, anchor=south, align=center}]
    \fill[black!7] (axis cs:0,-0.85) rectangle (axis cs:0.72,0.85);
    \draw[black!50, dashed, line width=0.6pt] (axis cs:0,-0.85) -- (axis cs:0,0.85);
    \draw[colorSig, line width=1.6pt] (axis cs:-0.0343,0) -- (axis cs:0.2721,0);
    \draw[colorSig, line width=1.6pt] (axis cs:-0.0343,-0.24) -- (axis cs:-0.0343,0.24);
    \draw[colorSig, line width=1.6pt] (axis cs:0.2721,-0.24) -- (axis cs:0.2721,0.24);
    \node[circle, fill=colorSig, inner sep=2.2pt] at (axis cs:0.1091,0) {};
  \nextgroupplot[ytick={0}, yticklabels={{iteration 1\\fixed}}, yticklabel style={align=right, font=\small}]
    \fill[black!7] (axis cs:0,-0.85) rectangle (axis cs:0.72,0.85);
    \draw[black!50, dashed, line width=0.6pt] (axis cs:0,-0.85) -- (axis cs:0,0.85);
    \draw[colorSig, line width=1.6pt] (axis cs:0.0398,0) -- (axis cs:0.2888,0);
    \draw[colorSig, line width=1.6pt] (axis cs:0.0398,-0.24) -- (axis cs:0.0398,0.24);
    \draw[colorSig, line width=1.6pt] (axis cs:0.2888,-0.24) -- (axis cs:0.2888,0.24);
    \node[circle, fill=colorSig, inner sep=2.2pt] at (axis cs:0.1539,0) {};
  \nextgroupplot[ytick=\empty]
    \fill[black!7] (axis cs:0,-0.85) rectangle (axis cs:0.72,0.85);
    \draw[black!50, dashed, line width=0.6pt] (axis cs:0,-0.85) -- (axis cs:0,0.85);
    \draw[colorSig, line width=1.6pt] (axis cs:-0.0320,0) -- (axis cs:0.2105,0);
    \draw[colorSig, line width=1.6pt] (axis cs:-0.0320,-0.24) -- (axis cs:-0.0320,0.24);
    \draw[colorSig, line width=1.6pt] (axis cs:0.2105,-0.24) -- (axis cs:0.2105,0.24);
    \node[circle, fill=colorSig, inner sep=2.2pt] at (axis cs:0.0779,0) {};
  \end{groupplot}
  \node[anchor=south west, font=\small] at ($(group c1r1.north west)+(-1.55cm,1.30cm)$) {\textbf{(b)} Post-training gain by condition};
  \node[anchor=north, font=\footnotesize, align=center] at ($(group c1r2.south)!0.5!(group c2r2.south)$) [yshift=-0.42cm] {Post-trained $-$ untrained base,\\GP-UCB-normalized endpoint, 0 = no change};
\end{tikzpicture}

%% file: figures/icbinb/tab_models.tex
\begin{table}[h]
  \centering\small
  \caption{The 19 evaluated models in release order, with the API slug used and the number of
  biology tasks contributing to each metric. Slugs are the exact routing identifiers recorded by
  the shared client registry. Six models were run through per-model evaluation harnesses that did
  not record the resolved slug, shown as ``--''.}
  \label{tab:models}
  \begin{tabular}{@{}llrr@{}}
    \toprule
    Model & API slug & Tasks (traj.) & Tasks (endp.) \\
    \midrule
    llama-3.1-8b & \texttt{meta-llama/llama-3.1-8b-instruct} & 44 & 44 \\
    deepseek-v3.2 & \texttt{deepseek/deepseek-v3.2} & 36 & 35 \\
    gemini-3-flash & \texttt{google/gemini-3-flash-preview} & 44 & 43 \\
    glm-4.7 & \texttt{z-ai/glm-4.7} & 45 & -- \\
    claude-sonnet-4.6 & \texttt{anthropic/claude-sonnet-4-6} & 42 & 41 \\
    gemini-3.1-pro & \texttt{google/gemini-3.1-pro-preview} & 45 & 42 \\
    gpt-5.4 & \texttt{openai/gpt-5.4} & 45 & 44 \\
    gpt-5.4-mini & \texttt{openai/gpt-5.4-mini} & 44 & 44 \\
    claude-opus-4.7 & \texttt{anthropic/claude-opus-4.7} & 36 & 34 \\
    gpt-5.5 & \texttt{openai/gpt-5.5} & 45 & 45 \\
    gemini-3.5-flash & \texttt{google/gemini-3.5-flash} & 45 & 45 \\
    claude-opus-4.8 & \texttt{anthropic/claude-opus-4.8} & 35 & 33 \\
    glm-5.2 & \texttt{z-ai/glm-5.2} & 45 & 44 \\
    gpt-5.6-sol & \texttt{--} & 45 & 45 \\
    gpt-5.6-terra & \texttt{--} & 45 & 45 \\
    muse-spark-1.1 & \texttt{--} & 21 & 21 \\
    kimi-k3 & \texttt{--} & 45 & 45 \\
    gemini-3.6-flash & \texttt{--} & 45 & 44 \\
    claude-opus-5 & \texttt{--} & 32 & 29 \\
    \bottomrule
  \end{tabular}
\end{table}

%% file: figures/icbinb/fig_models_vs_gp_env.tex
\begin{figure}[!htbp]
  \centering
  \input{figures/icbinb/fig_models_vs_gp}
  \caption{\textbf{No evaluated model outperforms GP-UCB by a meaningful margin, in either
  condition.} Median GP-UCB-normalized trajectory score for each of the 19 evaluated models, in
  release order, with 0 the GP-UCB reference. Circles are domain-agnostic, squares domain-aware, and
  the gray rule joins the two conditions for one model. Most models sit at or below the
  reference in both conditions and the largest positive margin is $+0.014$, while domain-agnostic is the higher of the two conditions on 14 of 19. No post-trained arms appear here.}
  \label{fig:models_vs_gp}
\end{figure}

%% file: figures/icbinb/fig_models_vs_gp.tex
\begin{tikzpicture}
  \begin{axis}[
    width=0.62\linewidth, height=8.2cm,
    xmin=-0.624, xmax=0.050, ymin=-0.8, ymax=18.8,
    xlabel={GP-UCB-normalized trajectory, median over tasks},
    xlabel style={font=\footnotesize},
    ytick={18,17,16,15,14,13,12,11,10,9,8,7,6,5,4,3,2,1,0},
    yticklabels={{llama-3.1-8b},{deepseek-v3.2},{gemini-3-flash},{glm-4.7},{claude-sonnet-4.6},{gemini-3.1-pro},{gpt-5.4},{gpt-5.4-mini},{claude-opus-4.7},{gpt-5.5},{gemini-3.5-flash},{claude-opus-4.8},{glm-5.2},{gpt-5.6-sol},{gpt-5.6-terra},{muse-spark-1.1},{kimi-k3},{gemini-3.6-flash},{claude-opus-5}},
    yticklabel style={font=\tiny}, xticklabel style={font=\scriptsize},
    axis lines*=left, tick style={black!60},
    xmajorgrids, grid style={black!12},
    legend style={font=\tiny, at={(0.02,-0.10)}, anchor=north west,
                  draw=none, fill=none, legend columns=2},
  ]
    \draw[black!55, dashed, line width=0.7pt] (axis cs:0,-0.8) -- (axis cs:0,18.8);
    \draw[black!25, line width=0.5pt] (axis cs:-0.5875,18) -- (axis cs:-0.1777,18);
    \draw[black!25, line width=0.5pt] (axis cs:-0.0779,17) -- (axis cs:-0.0205,17);
    \draw[black!25, line width=0.5pt] (axis cs:-0.5277,16) -- (axis cs:-0.0080,16);
    \draw[black!25, line width=0.5pt] (axis cs:-0.0083,15) -- (axis cs:-0.0019,15);
    \draw[black!25, line width=0.5pt] (axis cs:-0.0185,14) -- (axis cs:-0.0014,14);
    \draw[black!25, line width=0.5pt] (axis cs:-0.5045,13) -- (axis cs:0.0096,13);
    \draw[black!25, line width=0.5pt] (axis cs:-0.0604,12) -- (axis cs:-0.0038,12);
    \draw[black!25, line width=0.5pt] (axis cs:-0.1069,11) -- (axis cs:-0.0393,11);
    \draw[black!25, line width=0.5pt] (axis cs:-0.0028,10) -- (axis cs:0.0061,10);
    \draw[black!25, line width=0.5pt] (axis cs:-0.0258,9) -- (axis cs:-0.0053,9);
    \draw[black!25, line width=0.5pt] (axis cs:-0.0252,8) -- (axis cs:-0.0107,8);
    \draw[black!25, line width=0.5pt] (axis cs:0.0081,7) -- (axis cs:0.0142,7);
    \draw[black!25, line width=0.5pt] (axis cs:-0.0007,6) -- (axis cs:0.0034,6);
    \draw[black!25, line width=0.5pt] (axis cs:-0.0207,5) -- (axis cs:0.0007,5);
    \draw[black!25, line width=0.5pt] (axis cs:-0.0118,4) -- (axis cs:-0.0064,4);
    \draw[black!25, line width=0.5pt] (axis cs:-0.1461,3) -- (axis cs:-0.0026,3);
    \draw[black!25, line width=0.5pt] (axis cs:-0.0086,2) -- (axis cs:0.0050,2);
    \draw[black!25, line width=0.5pt] (axis cs:-0.0017,1) -- (axis cs:0.0008,1);
    \draw[black!25, line width=0.5pt] (axis cs:0.0090,0) -- (axis cs:0.0102,0);
    \addplot[only marks, mark=*, mark size=1.5pt, colorAgnostic] coordinates {(-0.1777,18) (-0.0205,17) (-0.0080,16) (-0.0019,15) (-0.0014,14) (0.0096,13) (-0.0038,12) (-0.0393,11) (-0.0028,10) (-0.0053,9) (-0.0252,8) (0.0081,7) (-0.0007,6) (0.0007,5) (-0.0064,4) (-0.0026,3) (-0.0086,2) (0.0008,1) (0.0102,0)};
    \addlegendentry{domain-agnostic}
    \addplot[only marks, mark=square*, mark size=1.4pt, colorContext] coordinates {(-0.5875,18) (-0.0779,17) (-0.5277,16) (-0.0083,15) (-0.0185,14) (-0.5045,13) (-0.0604,12) (-0.1069,11) (0.0061,10) (-0.0258,9) (-0.0107,8) (0.0142,7) (0.0034,6) (-0.0207,5) (-0.0118,4) (-0.1461,3) (0.0050,2) (-0.0017,1) (0.0090,0)};
    \addlegendentry{domain-aware}
  \end{axis}
\end{tikzpicture}

%% file: figures/icbinb/fig_two_metric_per_model_env.tex
\begin{figure}[!htbp]
  \centering
  \input{figures/icbinb/fig_two_metric_per_model}
  \caption{\textbf{No individual model shows a significant domain-aware advantage.} Per-model share
  of biology tasks on which domain-aware prompting outperforms domain-agnostic, under the trajectory
  metric (circles) and the endpoint metric (diamonds), for all 19 evaluated models in release
  order. Bars are Wilson 95\% confidence intervals and the solid rule is the 50\% no-effect null.
  Almost every per-model interval spans the null. Per-model tests are underpowered by design at
  four runs per pair, and primary inference uses the pooled and mixed-effects analyses
  (Fig.~\ref{fig:headline}). The three rows below the rule pool task-model pairs with a bootstrap
  over models, for comparison with the mixed-effects estimates. Those pooled rows are
  cell-weighted, so they differ from the mixed-effects values. We do not draw a release-cohort conclusion from this panel. The five July 2026 models were all scored through per-model evaluation summaries
  rather than the shared client registry (\S\ref{app:models}), so recency and scoring path are not
  separable here. Open markers identify those models. Four runs per task-condition pair.}
  \label{fig:two_metric}
\end{figure}

%% file: figures/icbinb/fig_two_metric_per_model.tex

\begin{tikzpicture}
  \begin{axis}[
    width=0.70\linewidth, height=11.36cm,
    xmin=0, xmax=100, ymin=-4.5, ymax=18.8,
    xlabel={Domain-aware win rate vs domain-agnostic (\%), biology},
    xlabel style={font=\footnotesize},
    ytick={-3.8,-2.7,-1.6,0,1,2,3,4,5,6,7,8,9,10,11,12,13,14,15,16,17,18}, yticklabels={\textbf{All 19 models},\textbf{Jul-26 cohort},\textbf{Dec-25--Jun-26},Llama 3.1 8B,DeepSeek V3.2,Gemini 3 Flash,GLM 4.7,Claude Sonnet 4.6,Gemini 3.1 Pro,GPT-5.4,GPT-5.4 Mini,Claude Opus 4.7,GPT-5.5,Gemini 3.5 Flash,Claude Opus 4.8,GLM 5.2,GPT-5.6 Sol,GPT-5.6 Terra,Muse Spark 1.1,Kimi K3,Gemini 3.6 Flash,Claude Opus 5},
    yticklabel style={font=\tiny}, xticklabel style={font=\scriptsize},
    axis lines*=left, tick style={black!60},
    grid=major, grid style={gray!20}, ymajorgrids=false,
    legend style={at={(0.99,0.02)}, anchor=south east, font=\tiny,
                  draw=black!25, fill=white, fill opacity=0.9,
                  text opacity=1, legend cell align=left},
  ]
    \draw[black!60, line width=1.1pt] (axis cs:50,-4.5) -- (axis cs:50,18.8);
    \node[font=\tiny, black!60, anchor=south, rotate=90] at (axis cs:50,18.9) {no effect (50\%)};
    \node[font=\tiny, black!45, anchor=west, rotate=90] at (axis cs:1.5,18.1) {newer models $\uparrow$};
    \draw[colorSig, line width=0.7pt, opacity=0.55] (axis cs:33.75,0.17) -- (axis cs:62.06,0.17);
    \node[circle, draw=colorSig, line width=0.6pt, inner sep=1.15pt, fill=colorSig] at (axis cs:47.73,0.17) {};
    \draw[colorSig, line width=0.9pt] (axis cs:33.75,-0.17) -- (axis cs:62.06,-0.17);
    \node[diamond, draw=colorSig, line width=0.6pt, inner sep=1.05pt, fill=colorSig] at (axis cs:47.73,-0.17) {};
    \draw[colorSig, line width=0.7pt, opacity=0.55] (axis cs:27.14,1.17) -- (axis cs:57.80,1.17);
    \node[circle, draw=colorSig, line width=0.6pt, inner sep=1.15pt, fill=colorSig] at (axis cs:41.67,1.17) {};
    \draw[colorSig, line width=0.9pt] (axis cs:32.99,0.83) -- (axis cs:64.43,0.83);
    \node[diamond, draw=colorSig, line width=0.6pt, inner sep=1.05pt, fill=colorSig] at (axis cs:48.57,0.83) {};
    \draw[colorSig, line width=0.7pt, opacity=0.55] (axis cs:25.72,2.17) -- (axis cs:53.38,2.17);
    \node[circle, draw=colorSig, line width=0.6pt, inner sep=1.15pt, fill=colorSig] at (axis cs:38.64,2.17) {};
    \draw[colorSig, line width=0.9pt] (axis cs:22.42,1.83) -- (axis cs:49.83,1.83);
    \node[diamond, draw=colorSig, line width=0.6pt, inner sep=1.05pt, fill=colorSig] at (axis cs:34.88,1.83) {};
    \draw[colorSig, line width=0.7pt, opacity=0.55] (axis cs:26.39,3.17) -- (axis cs:53.54,3.17);
    \node[circle, draw=colorSig, line width=0.6pt, inner sep=1.15pt, fill=colorSig, fill=white] at (axis cs:39.13,3.17) {};
    \draw[colorSig, line width=0.7pt, opacity=0.55] (axis cs:27.04,4.17) -- (axis cs:55.51,4.17);
    \node[circle, draw=colorSig, line width=0.6pt, inner sep=1.15pt, fill=colorSig] at (axis cs:40.48,4.17) {};
    \draw[colorSig, line width=0.9pt] (axis cs:27.76,3.83) -- (axis cs:56.63,3.83);
    \node[diamond, draw=colorSig, line width=0.6pt, inner sep=1.05pt, fill=colorSig] at (axis cs:41.46,3.83) {};
    \draw[colorSig, line width=0.7pt, opacity=0.55] (axis cs:21.36,5.17) -- (axis cs:47.93,5.17);
    \node[circle, draw=colorSig, line width=0.6pt, inner sep=1.15pt, fill=colorSig] at (axis cs:33.33,5.17) {};
    \draw[colorSig, line width=0.9pt] (axis cs:15.30,4.83) -- (axis cs:41.07,4.83);
    \node[diamond, draw=colorSig, line width=0.6pt, inner sep=1.05pt, fill=colorSig] at (axis cs:26.19,4.83) {};
    \draw[colorSig, line width=0.7pt, opacity=0.55] (axis cs:34.96,6.17) -- (axis cs:63.00,6.17);
    \node[circle, draw=colorSig, line width=0.6pt, inner sep=1.15pt, fill=colorSig] at (axis cs:48.89,6.17) {};
    \draw[colorSig, line width=0.9pt] (axis cs:20.00,5.83) -- (axis cs:46.56,5.83);
    \node[diamond, draw=colorSig, line width=0.6pt, inner sep=1.05pt, fill=colorSig] at (axis cs:31.82,5.83) {};
    \draw[colorSig, line width=0.7pt, opacity=0.55] (axis cs:25.72,7.17) -- (axis cs:53.38,7.17);
    \node[circle, draw=colorSig, line width=0.6pt, inner sep=1.15pt, fill=colorSig] at (axis cs:38.64,7.17) {};
    \draw[colorSig, line width=0.9pt] (axis cs:37.94,6.83) -- (axis cs:66.25,6.83);
    \node[diamond, draw=colorSig, line width=0.6pt, inner sep=1.05pt, fill=colorSig] at (axis cs:52.27,6.83) {};
    \draw[colorSig, line width=0.7pt, opacity=0.55] (axis cs:39.58,8.17) -- (axis cs:70.46,8.17);
    \node[circle, draw=colorSig, line width=0.6pt, inner sep=1.15pt, fill=colorSig] at (axis cs:55.56,8.17) {};
    \draw[colorSig, line width=0.9pt] (axis cs:39.45,7.83) -- (axis cs:71.12,7.83);
    \node[diamond, draw=colorSig, line width=0.6pt, inner sep=1.05pt, fill=colorSig] at (axis cs:55.88,7.83) {};
    \draw[colorSig, line width=0.7pt, opacity=0.55] (axis cs:34.12,9.17) -- (axis cs:61.86,9.17);
    \node[circle, draw=colorSig, line width=0.6pt, inner sep=1.15pt, fill=colorSig, fill=white] at (axis cs:47.83,9.17) {};
    \draw[colorSig, line width=0.9pt] (axis cs:25.11,8.83) -- (axis cs:52.37,8.83);
    \node[diamond, draw=colorSig, line width=0.6pt, inner sep=1.05pt, fill=colorSig, fill=white] at (axis cs:37.78,8.83) {};
    \draw[colorSig, line width=0.7pt, opacity=0.55] (axis cs:36.12,10.17) -- (axis cs:63.88,10.17);
    \node[circle, draw=colorSig, line width=0.6pt, inner sep=1.15pt, fill=colorSig, fill=white] at (axis cs:50.00,10.17) {};
    \draw[colorSig, line width=0.9pt] (axis cs:36.12,9.83) -- (axis cs:63.88,9.83);
    \node[diamond, draw=colorSig, line width=0.6pt, inner sep=1.05pt, fill=colorSig, fill=white] at (axis cs:50.00,9.83) {};
    \draw[colorSig, line width=0.7pt, opacity=0.55] (axis cs:49.15,11.17) -- (axis cs:79.17,11.17);
    \node[circle, draw=colorSig, line width=0.6pt, inner sep=1.15pt, fill=colorSig, fill=white] at (axis cs:65.71,11.17) {};
    \draw[colorSig, line width=0.9pt] (axis cs:29.84,10.83) -- (axis cs:62.01,10.83);
    \node[diamond, draw=colorSig, line width=0.6pt, inner sep=1.05pt, fill=colorSig, fill=white] at (axis cs:45.45,10.83) {};
    \draw[colorSig, line width=0.7pt, opacity=0.55] (axis cs:44.34,12.17) -- (axis cs:71.71,12.17);
    \node[circle, draw=colorSig, line width=0.6pt, inner sep=1.15pt, fill=colorSig, fill=white] at (axis cs:58.70,12.17) {};
    \draw[colorSig, line width=0.9pt] (axis cs:29.68,11.83) -- (axis cs:57.78,11.83);
    \node[diamond, draw=colorSig, line width=0.6pt, inner sep=1.05pt, fill=colorSig, fill=white] at (axis cs:43.18,11.83) {};
    \draw[colorSig, line width=0.7pt, opacity=0.55] (axis cs:30.21,13.17) -- (axis cs:57.75,13.17);
    \node[circle, draw=colorSig, line width=0.6pt, inner sep=1.15pt, fill=colorSig, fill=white] at (axis cs:43.48,13.17) {};
    \draw[colorSig, line width=0.9pt] (axis cs:22.68,12.83) -- (axis cs:49.23,12.83);
    \node[diamond, draw=colorSig, line width=0.6pt, inner sep=1.05pt, fill=colorSig, fill=white] at (axis cs:34.78,12.83) {};
    \draw[colorSig, line width=0.7pt, opacity=0.55] (axis cs:36.12,14.17) -- (axis cs:63.88,14.17);
    \node[circle, draw=colorSig, line width=0.6pt, inner sep=1.15pt, fill=colorSig, fill=white] at (axis cs:50.00,14.17) {};
    \draw[colorSig, line width=0.9pt] (axis cs:24.52,13.83) -- (axis cs:51.40,13.83);
    \node[diamond, draw=colorSig, line width=0.6pt, inner sep=1.05pt, fill=colorSig, fill=white] at (axis cs:36.96,13.83) {};
    \draw[colorSig, line width=0.7pt, opacity=0.55] (axis cs:13.81,15.17) -- (axis cs:49.96,15.17);
    \node[circle, draw=colorSig, line width=0.6pt, inner sep=1.15pt, fill=colorSig, fill=white] at (axis cs:28.57,15.17) {};
    \draw[colorSig, line width=0.9pt] (axis cs:10.63,14.83) -- (axis cs:45.09,14.83);
    \node[diamond, draw=colorSig, line width=0.6pt, inner sep=1.05pt, fill=colorSig, fill=white] at (axis cs:23.81,14.83) {};
    \draw[colorSig, line width=0.7pt, opacity=0.55] (axis cs:40.18,16.17) -- (axis cs:67.85,16.17);
    \node[circle, draw=colorSig, line width=0.6pt, inner sep=1.15pt, fill=colorSig, fill=white] at (axis cs:54.35,16.17) {};
    \draw[colorSig, line width=0.9pt] (axis cs:27.02,15.83) -- (axis cs:54.55,15.83);
    \node[diamond, draw=colorSig, line width=0.6pt, inner sep=1.05pt, fill=colorSig, fill=white] at (axis cs:40.00,15.83) {};
    \draw[colorSig, line width=0.7pt, opacity=0.55] (axis cs:30.21,17.17) -- (axis cs:57.75,17.17);
    \node[circle, draw=colorSig, line width=0.6pt, inner sep=1.15pt, fill=colorSig, fill=white] at (axis cs:43.48,17.17) {};
    \draw[colorSig, line width=0.9pt] (axis cs:27.69,16.83) -- (axis cs:55.59,16.83);
    \node[diamond, draw=colorSig, line width=0.6pt, inner sep=1.05pt, fill=colorSig, fill=white] at (axis cs:40.91,16.83) {};
    \draw[colorSig, line width=0.7pt, opacity=0.55] (axis cs:33.63,18.17) -- (axis cs:66.37,18.17);
    \node[circle, draw=colorSig, line width=0.6pt, inner sep=1.15pt, fill=colorSig, fill=white] at (axis cs:50.00,18.17) {};
    \draw[colorSig, line width=0.9pt] (axis cs:28.41,17.83) -- (axis cs:62.45,17.83);
    \node[diamond, draw=colorSig, line width=0.6pt, inner sep=1.05pt, fill=colorSig, fill=white] at (axis cs:44.83,17.83) {};
    \draw[colorSig, line width=2.1pt, opacity=0.55] (axis cs:41.18,-1.34) -- (axis cs:50.34,-1.34);
    \draw[colorSig, line width=2.1pt] (axis cs:41.18,-1.43) -- (axis cs:41.18,-1.25);
    \draw[colorSig, line width=2.1pt] (axis cs:50.34,-1.43) -- (axis cs:50.34,-1.25);
    \node[circle, fill=colorSig, fill opacity=0.55, draw=white, line width=0.5pt, inner sep=2.5pt] at (axis cs:45.66,-1.34) {};
    \node[font=\tiny, colorSig, anchor=west] at (axis cs:51.94,-1.34) {45.7\%};
    \draw[colorSig, line width=2.1pt, opacity=1.0] (axis cs:37.00,-1.86) -- (axis cs:46.55,-1.86);
    \draw[colorSig, line width=2.1pt] (axis cs:37.00,-1.95) -- (axis cs:37.00,-1.77);
    \draw[colorSig, line width=2.1pt] (axis cs:46.55,-1.95) -- (axis cs:46.55,-1.77);
    \node[diamond, fill=colorSig, fill opacity=1.0, draw=white, line width=0.5pt, inner sep=2.5pt] at (axis cs:41.86,-1.86) {};
    \node[font=\tiny, colorSig, anchor=west] at (axis cs:48.15,-1.86) {41.9\%};
    \draw[colorSig, line width=2.1pt, opacity=0.55] (axis cs:44.44,-2.44) -- (axis cs:52.13,-2.44);
    \draw[colorSig, line width=2.1pt] (axis cs:44.44,-2.53) -- (axis cs:44.44,-2.35);
    \draw[colorSig, line width=2.1pt] (axis cs:52.13,-2.53) -- (axis cs:52.13,-2.35);
    \node[circle, fill=colorSig, fill opacity=0.55, draw=white, line width=0.5pt, inner sep=2.5pt] at (axis cs:48.15,-2.44) {};
    \node[font=\tiny, colorSig, anchor=west] at (axis cs:53.73,-2.44) {48.1\%};
    \draw[colorSig, line width=2.1pt, opacity=1.0] (axis cs:36.52,-2.96) -- (axis cs:41.88,-2.96);
    \draw[colorSig, line width=2.1pt] (axis cs:36.52,-3.05) -- (axis cs:36.52,-2.87);
    \draw[colorSig, line width=2.1pt] (axis cs:41.88,-3.05) -- (axis cs:41.88,-2.87);
    \node[diamond, fill=colorSig, fill opacity=1.0, draw=white, line width=0.5pt, inner sep=2.5pt] at (axis cs:39.05,-2.96) {};
    \node[font=\tiny, colorSig, anchor=west] at (axis cs:43.48,-2.96) {39.0\%};
    \draw[colorSig, line width=2.1pt, opacity=0.55] (axis cs:42.80,-3.54) -- (axis cs:49.93,-3.54);
    \draw[colorSig, line width=2.1pt] (axis cs:42.80,-3.63) -- (axis cs:42.80,-3.45);
    \draw[colorSig, line width=2.1pt] (axis cs:49.93,-3.63) -- (axis cs:49.93,-3.45);
    \node[circle, fill=colorSig, fill opacity=0.55, draw=white, line width=0.5pt, inner sep=2.5pt] at (axis cs:46.34,-3.54) {};
    \node[font=\tiny, colorSig, anchor=west] at (axis cs:51.53,-3.54) {46.3\%};
    \draw[colorSig, line width=2.1pt, opacity=1.0] (axis cs:37.38,-4.06) -- (axis cs:44.65,-4.06);
    \draw[colorSig, line width=2.1pt] (axis cs:37.38,-4.15) -- (axis cs:37.38,-3.97);
    \draw[colorSig, line width=2.1pt] (axis cs:44.65,-4.15) -- (axis cs:44.65,-3.97);
    \node[diamond, fill=colorSig, fill opacity=1.0, draw=white, line width=0.5pt, inner sep=2.5pt] at (axis cs:41.05,-4.06) {};
    \node[font=\tiny, colorSig, anchor=west] at (axis cs:46.25,-4.06) {41.0\%};
    \draw[black!65, line width=0.9pt] (axis cs:0,-1.02) -- (axis cs:100,-1.02);
    \node[font=\scriptsize, black!65, anchor=west] at (axis cs:1,-1.28) {POOLED};
    \addlegendimage{colorSig, mark=*, only marks, mark options={fill opacity=0.55}}
    \addlegendentry{Trajectory (bsf-AUC@30)}
    \addlegendimage{colorSig, mark=diamond*, only marks}
    \addlegendentry{Endpoint (bsf-Outcome@30)}
  \end{axis}
\end{tikzpicture}

%% file: figures/icbinb/fig_reversal_env.tex
\begin{figure}[!htbp]
  \centering
  \input{figures/icbinb/fig_reversal}
  \caption{\textbf{Real parameter names help on education tasks but not on biology, so the null is not
  a general property of descriptive prompts.} Pooled rate at which domain-aware prompting outperforms
  domain-agnostic, under both metrics, on the biology panel (cobalt) and on the 10 education
  tasks (teal). Bars are 95\% bootstrap intervals resampling models. The dashed line is the 50\% no-effect null. On education both intervals sit well above the null, and on biology the endpoint interval lies below it. Counts at the right give the models and task-model pairs behind each row. Education endpoint rests on fewer models, since the later models' education runs do not all include the endpoint metric. Four runs per task-condition pair. Per-domain statistics in \S\ref{sec:eduspecific}.}
  \label{fig:reversal}
\end{figure}

%% file: figures/icbinb/fig_reversal.tex

\begin{tikzpicture}
  \begin{axis}[
    width=0.62\linewidth, height=4.08cm,
    xmin=15, xmax=118, ymin=-0.7, ymax=3.7,
    xtick={20,30,40,50,60,70,80,90},
    xlabel={Domain-aware win rate vs domain-agnostic (\%)},
    xlabel style={font=\footnotesize},
    ytick={0,1,2,3},
    yticklabels={{Trajectory: biology},{Trajectory: education},{Endpoint: biology},{Endpoint: education}},
    yticklabel style={font=\tiny}, xticklabel style={font=\scriptsize},
    axis lines*=left, tick style={black!60},
    grid=major, grid style={gray!20}, ymajorgrids=false,
  ]
    \draw[black!45, dashed, thick] (axis cs:50,-0.7) -- (axis cs:50,3.7);
    \node[font=\tiny, black!55, anchor=south east, rotate=90] at (axis cs:50,3.9) {no effect (50\%)};
    \draw[colorSig, line width=1.3pt] (axis cs:42.69,0) -- (axis cs:50.00,0);
    \node[circle, fill=colorSig, inner sep=1.7pt] at (axis cs:46.34,0) {};
    \node[font=\tiny, black!55, anchor=west] at (axis cs:51.50,0) {19 models, 792 pairs};
    \draw[colorEdu, line width=1.3pt] (axis cs:61.73,1) -- (axis cs:74.72,1);
    \node[circle, fill=colorEdu, inner sep=1.7pt] at (axis cs:68.48,1) {};
    \node[font=\tiny, black!55, anchor=west] at (axis cs:76.22,1) {19 models, 165 pairs};
    \draw[colorSig, line width=1.3pt] (axis cs:37.50,2) -- (axis cs:44.51,2);
    \node[circle, fill=colorSig, inner sep=1.7pt] at (axis cs:41.05,2) {};
    \node[font=\tiny, black!55, anchor=west] at (axis cs:46.01,2) {18 models, 726 pairs};
    \draw[colorEdu, line width=1.3pt] (axis cs:55.38,3) -- (axis cs:81.08,3);
    \node[circle, fill=colorEdu, inner sep=1.7pt] at (axis cs:68.12,3) {};
    \node[font=\tiny, black!55, anchor=west] at (axis cs:82.58,3) {8 models, 69 pairs};
  \end{axis}
\end{tikzpicture}

%% file: figures/icbinb/tab_divergent6.tex
\begin{table}[t]
  \centering
  \small
  \caption{\textbf{The six literature-divergent tasks.} On each of these tasks, the categorical value used most often across the source papers is not the value used by the single best-performing published experiment. Domain-agnostic prompting proposes that best-experiment value more often than domain-aware on all six, a paired mean of $+18.1$ percentage points. Two rows carry data caveats. The CHL-YN convention is a 3--3 modal tie and the transaminase conversion column has values above 100, and dropping either or both strengthens the effect to $+20.0$, $+19.5$ and $+22.2$ percentage points respectively, positive on every task in every variant (\S\ref{sec:anchoring}). Source DOIs per task in \S\ref{app:divergent}.}
  \label{tab:divergent6}
  \begin{tabular}{llll}
    \toprule
    Task & Key categorical & Literature convention & Published best \\
    \midrule
    HEK293 prime editing & delivery method & Transfection & Baculovirus \\
    Microalgae biomass & reactor type & Photobioreactor & Open raceway pond \\
    VHH nanobody expression & expression vector & pET-22b & pCold-I \\
    CHL-YN cell line & feed-rate strategy & Dynamic & Adaptive \\
    CHO antibody titer & CHO cell line & CHO-S & CHO-K1 \\
    Transaminase amination & cosolvent & Toluene & ChCl:EG:1,2-PD (deep eutectic) \\
    \bottomrule
  \end{tabular}
\end{table}

%% file: figures/icbinb/fig_anchoring_permodel_env.tex
\begin{figure}[!htbp]
  \centering
  \input{figures/icbinb/fig_anchoring_permodel}
  \caption{\textbf{The anchoring effect holds model by model, so no single model produces it.}
  Each arrow runs from a model's domain-agnostic rate (open circle) to its
  domain-aware rate on the six pairs where convention and the literature optimum diverge. Heavier arrows are models released after March 2026. The shaded region is everything below GP-UCB, an optimizer given
  no biological context at all. The three-way ordering is descriptive, since neither pairwise gap reaches
  significance. The direction is shared across nearly every model, so no single model drives the pooled result. }
  \label{fig:anchoring_permodel}
\end{figure}

%% file: figures/icbinb/fig_anchoring_permodel.tex
\begin{tikzpicture}
  \begin{axis}[
    width=0.42\linewidth, height=5.0cm,
    xmin=0, xmax=56.0, ymin=-0.8, ymax=11.8,
    xlabel={Match to published-best category (\%)},
    xlabel style={font=\small},
    ytick={0,1,2,3,4,5,6,7,8,9,10,11},
    yticklabels={Llama 3.1 8B,GPT-5.6 Sol,Opus 5,Gemini 3.1 Pro,Gemini 3.6 Flash,Kimi K3,Opus 4.7,Gemini 3 Flash,GPT-5.4 Mini,GPT-5.4,DeepSeek V3.2,Sonnet 4.6},
    yticklabel style={font=\tiny}, xticklabel style={font=\scriptsize},
    axis lines*=left, tick style={black!60},
    grid=major, grid style={gray!20}, ymajorgrids=false,
    title={Per model, six divergent pairs},
    title style={font=\footnotesize, at={(0,1.03)}, anchor=south west},
  ]
    \fill[black!6] (axis cs:0,-0.8) rectangle (axis cs:29.2,11.8);
    \draw[black!70, line width=1.0pt] (axis cs:29.2,-0.8) -- (axis cs:29.2,11.8);
    \node[font=\tiny, black!70, anchor=south] at (axis cs:29.2,11.7) {GP-UCB (29.2\%)};
    \draw[->, colorContext, line width=0.5pt] (axis cs:33.33,0) -- (axis cs:7.14,0);
    \node[circle, draw=colorAgnostic, fill=white, line width=0.5pt, inner sep=0.9pt] at (axis cs:33.33,0) {};
    \draw[->, colorContext, line width=0.9pt] (axis cs:37.50,1) -- (axis cs:12.50,1);
    \node[circle, draw=colorAgnostic, fill=white, line width=0.5pt, inner sep=0.9pt] at (axis cs:37.50,1) {};
    \draw[->, colorContext, line width=0.9pt] (axis cs:37.50,2) -- (axis cs:16.67,2);
    \node[circle, draw=colorAgnostic, fill=white, line width=0.5pt, inner sep=0.9pt] at (axis cs:37.50,2) {};
    \draw[->, colorContext, line width=0.5pt] (axis cs:43.48,3) -- (axis cs:16.67,3);
    \node[circle, draw=colorAgnostic, fill=white, line width=0.5pt, inner sep=0.9pt] at (axis cs:43.48,3) {};
    \draw[->, colorContext, line width=0.9pt] (axis cs:45.83,4) -- (axis cs:16.67,4);
    \node[circle, draw=colorAgnostic, fill=white, line width=0.5pt, inner sep=0.9pt] at (axis cs:45.83,4) {};
    \draw[->, colorContext, line width=0.9pt] (axis cs:41.67,5) -- (axis cs:20.83,5);
    \node[circle, draw=colorAgnostic, fill=white, line width=0.5pt, inner sep=0.9pt] at (axis cs:41.67,5) {};
    \draw[->, colorContext, line width=0.5pt] (axis cs:29.17,6) -- (axis cs:25.00,6);
    \node[circle, draw=colorAgnostic, fill=white, line width=0.5pt, inner sep=0.9pt] at (axis cs:29.17,6) {};
    \draw[->, colorContext, line width=0.5pt] (axis cs:20.83,7) -- (axis cs:25.00,7);
    \node[circle, draw=colorAgnostic, fill=white, line width=0.5pt, inner sep=0.9pt] at (axis cs:20.83,7) {};
    \draw[->, colorContext, line width=0.5pt] (axis cs:25.00,8) -- (axis cs:25.00,8);
    \node[circle, draw=colorAgnostic, fill=white, line width=0.5pt, inner sep=0.9pt] at (axis cs:25.00,8) {};
    \draw[->, colorContext, line width=0.5pt] (axis cs:45.83,9) -- (axis cs:26.09,9);
    \node[circle, draw=colorAgnostic, fill=white, line width=0.5pt, inner sep=0.9pt] at (axis cs:45.83,9) {};
    \draw[->, colorContext, line width=0.5pt] (axis cs:50.00,10) -- (axis cs:31.25,10);
    \node[circle, draw=colorAgnostic, fill=white, line width=0.5pt, inner sep=0.9pt] at (axis cs:50.00,10) {};
    \draw[->, colorContext, line width=0.5pt] (axis cs:45.83,11) -- (axis cs:43.75,11);
    \node[circle, draw=colorAgnostic, fill=white, line width=0.5pt, inner sep=0.9pt] at (axis cs:45.83,11) {};
  \end{axis}
\end{tikzpicture}

%% file: figures/icbinb/fig_optimizer_comparison_env.tex
\begin{figure}[!htbp]
  \centering
  \input{figures/icbinb/fig_optimizer_comparison}
  \caption{\textbf{Direct preference optimization against GP-UCB's choices lifts an 8B model to parity with it and to
  frontier-model performance, but only when the parameter names are not biologically meaningful.}
  GP-UCB-normalized trajectory score on 16 held-out biology tasks. Circles are medians over
  tasks, bars 95\% bootstrap intervals on the median. Zero is GP-UCB parity in \emph{(a)} and
  \emph{(b)}, and no difference between arms in \emph{(c)}. \emph{(a)} Under domain-agnostic prompts the DPO-trained model reaches parity with an interval an order of
  magnitude tighter than either other arm. \emph{(b)} Under domain-aware prompts the same three optimizers all fall well below parity and the
  intervals overlap completely, because task-to-task difficulty in this condition is far larger
  than the differences between arms. \emph{(c)} Differencing within task removes that
  difficulty and is where the comparison can actually be read. DPO is worth $+0.40$
  under domain-agnostic prompts (13 of 16 tasks) but $+0.03$ under domain-aware prompts, so the
  gain does not transfer.
  Claude and the DPO-trained 8B are indistinguishable in both conditions. Counts at the right are
  tasks favoring the first arm. $^{\dagger}$averaged over four training seeds, four runs per task each. The other arms are single models with four runs per task. One task
  (\texttt{baculovirus\_titer\_sf9}) is excluded, where a near-zero GP denominator sends the normalized score to $+32.8$ (see \S\ref{app:optimizers}).}
  \label{fig:optimizer_comparison}
\end{figure}

%% file: figures/icbinb/fig_optimizer_comparison.tex
\begin{tikzpicture}
  \begin{axis}[
    width=0.60\linewidth, height=2.62cm,
    xmin=-0.84, xmax=0.21, ymin=-0.7, ymax=2.70,
    xlabel={GP-UCB-normalized trajectory (0 = parity)},
    xlabel style={font=\footnotesize},
    ytick={0,1,2}, yticklabels={Llama 3.1 8B,Llama 3.1 8B + DPO vs GP-UCB$^{\dagger}$,Claude Sonnet 4.6},
    yticklabel style={font=\tiny}, xticklabel style={font=\scriptsize},
    axis lines*=left, tick style={black!60},
    scaled x ticks=false, xticklabel style={/pgf/number format/fixed, /pgf/number format/precision=2, font=\scriptsize},
    grid=major, grid style={gray!20}, ymajorgrids=false,
    title={\textbf{(a)} Domain-agnostic}, title style={font=\footnotesize, at={(0,1.02)}, anchor=south west},
  ]
    \draw[black!55, dashed] (axis cs:0,-0.7) -- (axis cs:0,2.70) node[above, font=\tiny, black!55] {GP-UCB};
    \draw[black!60, line width=1.2pt] (axis cs:-0.5916,0) -- (axis cs:-0.0556,0);
    \node[circle, fill=black!60, inner sep=1.6pt] at (axis cs:-0.3590,0) {};
    \draw[colorContext, line width=1.2pt] (axis cs:-0.0811,1) -- (axis cs:0.0390,1);
    \node[circle, fill=colorContext, inner sep=1.6pt] at (axis cs:-0.0141,1) {};
    \draw[colorAgnostic, line width=1.2pt] (axis cs:-0.4630,2) -- (axis cs:0.1645,2);
    \node[circle, fill=colorAgnostic, inner sep=1.6pt] at (axis cs:-0.0077,2) {};
  \end{axis}
  \begin{scope}[yshift=-3.4cm]
  \begin{axis}[
    width=0.60\linewidth, height=2.62cm,
    xmin=-0.84, xmax=0.21, ymin=-0.7, ymax=2.70,
    xlabel={GP-UCB-normalized trajectory (0 = parity)},
    xlabel style={font=\footnotesize},
    ytick={0,1,2}, yticklabels={Llama 3.1 8B,Llama 3.1 8B + DPO vs GP-UCB$^{\dagger}$,Claude Sonnet 4.6},
    yticklabel style={font=\tiny}, xticklabel style={font=\scriptsize},
    axis lines*=left, tick style={black!60},
    scaled x ticks=false, xticklabel style={/pgf/number format/fixed, /pgf/number format/precision=2, font=\scriptsize},
    grid=major, grid style={gray!20}, ymajorgrids=false,
    title={\textbf{(b)} Domain-aware}, title style={font=\footnotesize, at={(0,1.02)}, anchor=south west},
  ]
    \draw[black!55, dashed] (axis cs:0,-0.7) -- (axis cs:0,2.70) node[above, font=\tiny, black!55] {GP-UCB};
    \draw[black!60, line width=1.2pt] (axis cs:-0.7897,0) -- (axis cs:0.0063,0);
    \node[circle, fill=black!60, inner sep=1.6pt] at (axis cs:-0.3003,0) {};
    \draw[colorContext, line width=1.2pt] (axis cs:-0.7875,1) -- (axis cs:0.0629,1);
    \node[circle, fill=colorContext, inner sep=1.6pt] at (axis cs:-0.2316,1) {};
    \draw[colorAgnostic, line width=1.2pt] (axis cs:-0.7233,2) -- (axis cs:0.0674,2);
    \node[circle, fill=colorAgnostic, inner sep=1.6pt] at (axis cs:-0.2375,2) {};
  \end{axis}
  \end{scope}
  \begin{scope}[yshift=-7.2cm]
  \begin{axis}[
    width=0.60\linewidth, height=3.06cm,
    xmin=-0.38, xmax=0.86, ymin=-0.7, ymax=3.70,
    xlabel={Paired difference in GP-UCB-normalized trajectory},
    xlabel style={font=\footnotesize},
    ytick={0,1,2,3},
    yticklabels={{DPO $-$ base, domain-agnostic},{DPO $-$ base, domain-aware},{Claude $-$ DPO, domain-agnostic},{Claude $-$ DPO, domain-aware}},
    yticklabel style={font=\tiny}, xticklabel style={font=\scriptsize},
    axis lines*=left, tick style={black!60},
    scaled x ticks=false, xticklabel style={/pgf/number format/fixed, /pgf/number format/precision=2, font=\scriptsize},
    grid=major, grid style={gray!20}, ymajorgrids=false,
    title={\textbf{(c)} Paired within task}, title style={font=\footnotesize, at={(0,1.02)}, anchor=south west},
  ]
    \draw[black!55, dashed] (axis cs:0,-0.7) -- (axis cs:0,3.70) node[above, font=\tiny, black!55] {no difference};
    \draw[colorAgnostic, line width=1.2pt] (axis cs:0.1931,0) -- (axis cs:0.6391,0);
    \node[circle, fill=colorAgnostic, inner sep=1.6pt] at (axis cs:0.4002,0) {};
    \node[font=\tiny, black!55, anchor=west] at (axis cs:0.6691,0) {13/16};
    \draw[black!45, line width=1.2pt] (axis cs:-0.1070,1) -- (axis cs:0.1588,1);
    \node[circle, fill=black!45, inner sep=1.6pt] at (axis cs:0.0261,1) {};
    \node[font=\tiny, black!55, anchor=west] at (axis cs:0.1888,1) {10/16};
    \draw[black!45, line width=1.2pt] (axis cs:-0.2791,2) -- (axis cs:0.0668,2);
    \node[circle, fill=black!45, inner sep=1.6pt] at (axis cs:-0.0918,2) {};
    \node[font=\tiny, black!55, anchor=west] at (axis cs:0.0968,2) {8/15};
    \draw[black!45, line width=1.2pt] (axis cs:-0.0981,3) -- (axis cs:0.0712,3);
    \node[circle, fill=black!45, inner sep=1.6pt] at (axis cs:-0.0177,3) {};
    \node[font=\tiny, black!55, anchor=west] at (axis cs:0.1012,3) {6/14};
  \end{axis}
  \end{scope}
\end{tikzpicture}

%% file: figures/icbinb/fig_matched_prior_env.tex
\begin{figure}[!htbp]
  \centering
  \input{figures/icbinb/fig_matched_prior}
  \caption{\textbf{Most of what looks like a frontier optimization advantage is the opening
  proposal, and the four arms largely converge once every optimizer is handed the same first
  design.} Every arm
  receives one identical Claude Sonnet 4.6 domain-aware iteration-1 design per task, injected at
  iteration 1, and then optimizes iterations 2--30 itself under domain-agnostic prompts, on nine held-out biology tasks. \emph{(a)} One faint dot per task, with the arm's median marked. The dotted line is GP-UCB given the same opening. The four arms span $0.11$ in median, compared with a spread more than twice that when each chooses its own start, and the untrained 8B moves from
  well below Claude to within $0.004$ of it. \emph{(b)} The paired differences this design
  permits, one circle per task with the mean, a 95\% bootstrap interval, and a one-sided
  Wilcoxon. Differencing within
  task restores the ordering that the medians hide. Claude still outperforms the untrained base
  ($+0.112$, 7 of 9), which is the power check, while Claude and the DPO-trained 8B remain
  indistinguishable. The star marks the same Claude model given the same
  iteration-1 design but allowed to keep the biological names for iterations 2--30. That arm's median,
  $-0.107$, is indistinguishable from the $-0.110$ the same model scores under domain-agnostic
  prompts. With the
  prior held fixed the prompt effect is not visible between tasks and appears only when
  differenced within them, which is what Appendix~\ref{app:matched} does. Two arms are served
  locally and two through the API, so the Claude-versus-base contrast is the like-for-like one.
  Per-task values in \S\ref{app:matched}.}
  \label{fig:matched_prior}
\end{figure}

%% file: figures/icbinb/fig_matched_prior.tex
\begin{tikzpicture}
  \begin{axis}[
    width=0.62\linewidth, height=3.6cm,
    xmin=-0.99, xmax=0.36, ymin=-0.8, ymax=2.8,
    xlabel={GP-UCB-normalized trajectory (0 = parity)},
    xlabel style={font=\footnotesize},
    ytick={0,1,2},
    yticklabels={{Llama 3.1 8B untrained},{Claude Sonnet 4.6},{Llama 3.1 8B + DPO vs GP-UCB}},
    yticklabel style={font=\tiny}, xticklabel style={font=\scriptsize},
    axis lines*=left, tick style={black!60},
    grid=major, grid style={gray!20}, ymajorgrids=false,
    title={\textbf{(a)} Same opening design, then each optimizer searches},
    title style={font=\footnotesize, at={(0,1.02)}, anchor=south west},
  ]
    \draw[black!55, dashed] (axis cs:0,-0.8) -- (axis cs:0,2.8) node[above, font=\tiny, black!55] {GP-UCB parity};
    \addplot[only marks, mark=*, mark size=0.9pt, black!25] coordinates {(-0.6541,0.16) (-0.3144,0.16) (-0.0508,0.16) (-0.6493,0.16) (-0.1135,0.16) (-0.8726,0.16) (0.0949,0.16) (0.1662,0.16) (0.1204,0.16)};
    \draw[black!55, line width=1.6pt] (axis cs:-0.1135,-0.20) -- (axis cs:-0.1135,0.12);
    \node[font=\tiny, black!60, anchor=west] at (axis cs:0.2027,0.00) {-0.114};
    \addplot[only marks, mark=*, mark size=0.9pt, black!25] coordinates {(-0.1792,1.16) (-0.3114,1.16) (-0.0662,1.16) (-0.3117,1.16) (-0.1102,1.16) (-0.8259,1.16) (0.2534,1.16) (0.1641,1.16) (0.1211,1.16)};
    \draw[colorAgnostic, line width=1.6pt] (axis cs:-0.1102,0.80) -- (axis cs:-0.1102,1.12);
    \node[font=\tiny, black!60, anchor=west] at (axis cs:0.2027,1.00) {-0.110};
    \addplot[only marks, mark=*, mark size=0.9pt, black!25] coordinates {(0.0316,2.16) (-0.0957,2.16) (-0.0521,2.16) (-0.1986,2.16) (-0.1377,2.16) (-0.7131,2.16) (0.1217,2.16) (0.0496,2.16) (0.1204,2.16)};
    \draw[colorContext, line width=1.6pt] (axis cs:-0.0521,1.80) -- (axis cs:-0.0521,2.12);
    \node[font=\tiny, black!60, anchor=west] at (axis cs:0.2027,2.00) {-0.052};
    \draw[black!70, line width=0.8pt, dotted] (axis cs:-0.0063,-0.8) -- (axis cs:-0.0063,2.8);
    \node[star, star points=5, fill=colorContext, inner sep=1.3pt] at (axis cs:-0.1066,1) {};
    \node[font=\tiny, colorContext, anchor=north] at (axis cs:-0.1066,0.80) {same prior, keeps the names: -0.107};
  \end{axis}
  \begin{scope}[yshift=-4.0cm]
  \begin{axis}[
    width=0.62\linewidth, height=2.9cm,
    xmin=-0.19, xmax=0.93, ymin=-0.8, ymax=2.8,
    xlabel={Paired difference, same task},
    xlabel style={font=\footnotesize},
    ytick={0,1,2},
    yticklabels={{Claude $-$ untrained base},{DPO vs GP-UCB $-$ untrained base},{DPO vs GP-UCB $-$ Claude}},
    yticklabel style={font=\tiny}, xticklabel style={font=\scriptsize},
    axis lines*=left, tick style={black!60},
    grid=major, grid style={gray!20}, ymajorgrids=false,
    title={\textbf{(b)} Paired within task},
    title style={font=\footnotesize, at={(0,1.02)}, anchor=south west},
  ]
    \draw[black!55, dashed] (axis cs:0,-0.8) -- (axis cs:0,2.8) node[above, font=\tiny, black!55] {no difference};
    \addplot[only marks, mark=o, mark size=1.1pt, black!35] coordinates {(0.4749,0.20) (0.0030,0.20) (-0.0154,0.20) (0.3376,0.20) (0.0033,0.20) (0.0468,0.20) (0.1585,0.20) (-0.0021,0.20) (0.0007,0.20)};
    \draw[colorAgnostic, line width=1.2pt] (axis cs:0.0127,-0.16) -- (axis cs:0.2346,-0.16);
    \node[circle, fill=colorAgnostic, inner sep=1.5pt] at (axis cs:0.1119,-0.16) {};
    \node[font=\tiny, black!60, anchor=east] at (axis cs:0.9157,-0.16) {7/9, $p=0.04$};
    \addplot[only marks, mark=o, mark size=1.1pt, black!35] coordinates {(0.6857,1.20) (0.2187,1.20) (-0.0013,1.20) (0.4507,1.20) (-0.0242,1.20) (0.1596,1.20) (0.0268,1.20) (-0.1165,1.20) (0.0000,1.20)};
    \draw[colorAgnostic, line width=1.2pt] (axis cs:0.0109,0.84) -- (axis cs:0.3359,0.84);
    \node[circle, fill=colorAgnostic, inner sep=1.5pt] at (axis cs:0.1555,0.84) {};
    \node[font=\tiny, black!60, anchor=east] at (axis cs:0.9157,0.84) {6/9, $p=0.08$};
    \addplot[only marks, mark=o, mark size=1.1pt, black!35] coordinates {(0.2108,2.20) (0.2157,2.20) (0.0141,2.20) (0.1131,2.20) (-0.0275,2.20) (0.1128,2.20) (-0.1317,2.20) (-0.1145,2.20) (-0.0007,2.20)};
    \draw[black!45, line width=1.2pt] (axis cs:-0.0339,1.84) -- (axis cs:0.1218,1.84);
    \node[circle, fill=black!45, inner sep=1.5pt] at (axis cs:0.0436,1.84) {};
    \node[font=\tiny, black!60, anchor=east] at (axis cs:0.9157,1.84) {5/9, $p=0.29$};
  \end{axis}
  \end{scope}
\end{tikzpicture}